\pdfoutput=1

\documentclass[twocolumn]{article}

\usepackage[total={7.2in, 9in}]{geometry}

\usepackage{graphicx}
\usepackage{caption}
\usepackage{subcaption}
\usepackage{xcolor,colortbl}
\usepackage{amsopn}
\usepackage{amsmath}
\usepackage{amssymb}
\usepackage{xspace}
\usepackage{comment}
\usepackage{amsfonts,amssymb}

\usepackage{algorithm}
\usepackage[noend]{algpseudocode}

\usepackage{afterpage}
\newlength{\oldtextfloatsep}

\newcommand{\trans}{^{\!\mathsf{T}}}

\newcommand{\Laplace}{\mathop{}\!\mathbin\bigtriangleup}
\renewcommand{\div}{\nabla\!\cdot\!}
\newcommand{\RNum}[1]{{\bf (\lowercase\expandafter{\romannumeral #1\relax})}}
\newcommand{\dx}{\mathrm{d}x}

\newcommand{\wrt}{w.r.t.\@\xspace}
\newcommand{\cf}{cf.\@\xspace}
\newcommand{\Ie}{I.e.\@\xspace}
\newcommand{\ie}{i.e.\@\xspace}

\newcommand{\eg}{e.g.\@\xspace}
\newcommand{\etal}{et al.\@\xspace}
\newcommand{\Alg}{Alg.\@\xspace}
\newcommand{\Eq}{Eq.\@\xspace}

\newcommand{\Sec}{Sec.\@\xspace}

\DeclareMathOperator*{\argmin}{arg\,min}

\DeclareMathOperator*{\ell0}{l_0}

\definecolor{myRed}{rgb}{0.6,0.1,0.}
\definecolor{myBlue}{rgb}{0.0,0.2,0.6}
\definecolor{myGreen}{rgb}{0.2,0.6,0.}

\newcommand{\invisible}[1]{}

\newcommand{\Fref}[1]{Figure\xspace\ref{#1}\xspace}
\newcommand{\fref}[1]{Figure\xspace\ref{#1}\xspace}
\newcommand{\tref}[1]{Table\xspace\ref{#1}\xspace}
\newcommand{\sref}[1]{Section\xspace\ref{#1}\xspace}
\newcommand{\eref}[1]{\xspace\ref{#1}\xspace}

\begin{document}

\title{Variational 3D-PIV with Sparse Descriptors}

\author{Katrin Lasinger$^1$, Christoph Vogel$^2$, Thomas Pock$^{2,3}$ and Konrad Schindler$^1$
\vspace{5pt}
\\
$^1$Photogrammetry and Remote Sensing, ETH Zurich, Switzerland \\
$^2$Institute of Computer Graphics \& Vision, TU Graz, Austria \\
$^3$Austrian Institute of Technology, Austria
}
\date{}

\twocolumn[
 \begin{@twocolumnfalse}
 \maketitle

\begin{abstract}
\normalsize
3D Particle Imaging Velocimetry (3D-PIV) aim to recover the flow field
in a volume of fluid, which has been seeded with tracer particles and
observed from multiple camera viewpoints.

The first step of 3D-PIV is to reconstruct the 3D locations of the
tracer particles from synchronous views of the volume.
We propose a new method for iterative particle reconstruction (IPR),
in which the locations and intensities of all particles are inferred
in one joint energy minimization. The energy function is designed to
penalize deviations between the reconstructed 3D particles and the
image evidence, while at the same time aiming for a sparse set of
particles.
We find that the new method, without any post-processing, achieves
significantly cleaner particle volumes than a conventional,
tomographic MART reconstruction, and can handle a wide range of
particle densities.

The second step of 3D-PIV is to then recover the dense motion field
from two consecutive particle reconstructions.
We propose a variational model, which makes it possible to directly
include physical properties, such as incompressibility and viscosity,
in the estimation of the motion field.
To further exploit the sparse nature of the input data, we propose a
novel, compact descriptor of the local particle layout.
Hence, we avoid the memory-intensive storage of high-resolution
intensity volumes. Our framework is generic and allows for a variety
of different data costs (correlation measures) and regularizers. We
quantitatively evaluate it with both the sum of squared differences
(SSD) and the normalized cross-correlation (NCC), respectively with
both a hard and a soft version of the incompressibility constraint.
\end{abstract}
 \end{@twocolumnfalse}
 
\vspace{10pt}
]

\section{Introduction}
Experimental fluid dynamics requires a measurement system to observe
the flow in a fluid.
The main method to reconstruct the dense flow field in a (transparent)
volume of fluid is to inject neutral-buoyancy tracer particles, tune
the illumination to obtain high contrast at particle silhouettes, and
record their motion with multiple high-speed cameras.
The flow is then reconstructed either by \emph{Particle Tracking
Velocimetry}~(PTV)~\cite{kaz-00}, meaning that one follows the
trajectories of individual particles and interpolates the motion in
the rest of the volume; or by \emph{Particle Imaging
Velocimetry}~(PIV)~\cite{adr-11,raf-13}, where one directly estimates
the local motion everywhere in the volume from the
displacements of nearby particle constellations.

Early approaches operated essentially in 2D, focusing on a thin
slice of the volume that is selectively illuminated with a
laser light-sheet.
More recently, it has become common practice to process entire volumes
and recover actual 3D fluid motion.
A popular variant is to directly lift the problem to 3D by
discretizing the volumetric domain into voxels.
The standard approach is to process the multi-view video in two
steps.
First, a 3D particle distribution is reconstructed independently
for each frame, often by means of \emph{tomographic} PIV (Tomo-PIV)
\cite{els-06,atk-09,dis-12,cha-14}, and stored as an intensity/probability
volume, representing the particle-in-voxel likelihood.
A second step then recovers motion vectors, by densely matching the
reconstructed 3D particles (respectively, particle likelihoods)
between consecutive time frames.
The most widely used approach for the latter step is local matching of
relatively large interrogation volumes, followed by various
post-processing steps, \eg~\cite{els-06,Champagnat2011,yegavian-16}.
In computer vision terminology, this corresponds to the classic
Lucas-Kanade (LK) scheme for optical flow computation~\cite{lucas-81}.
While LK is no longer state-of-the-art for 2D flow, it is still
attractive in the 3D setting, due to its simplicity, and to the
efficiency inherent in a purely local model that lends itself to
parallel computation.
Optical flow, i.e., dense reconstruction of apparent motion in the 2D
image plane, is one of the most deeply studied problems in
computer vision.
Countless variants have been developed and have found their way into
practical applications, such as medical image registration, human
motion analysis and driver assistance.
The seminal work of Horn and Schunk~\cite{hor-81}, and its
contemporary descendants, \eg~\cite{bro-04,zac-07,wer-10}, address the
problem from a variational perspective. A setting we also
follow here.
Flow estimation from images is locally underconstrained
(\emph{aperture problem}).
Imperfect particle reconstructions and the ambiguity between
different, visually indistinguishable, particles further complicate
the task.
The normal solution for such an ill-posed, inverse problem is to
introduce prior knowledge that further constrains the result and acts
as a regularizer.

Indeed, modeling the problem in the volumetric domain opens up the
possibility to incorporate physical constraints into the model,
e.g., one can impose incompressibility of the fluid.
These advantages were first pointed out in \cite{hei-09}.
\cite{alv-09} were among the first to exploit physical properties
of fluids, however, only as a post-process to refine an initial
solution obtained with taking into account the physics.
Following our own previous work \cite{las-17}, we base regularization
on the specific properties of liquids and propose an energy, whose
optimality conditions correspond to the stationary Stokes equations of
fluid dynamics.
This strictly enforces a divergence-free flow field and additionally
penalizes the squared gradient of the flow, making explicit the
viscosity of the fluid.

\begin{figure}[tb]
 \centering
 \includegraphics[width=\columnwidth]{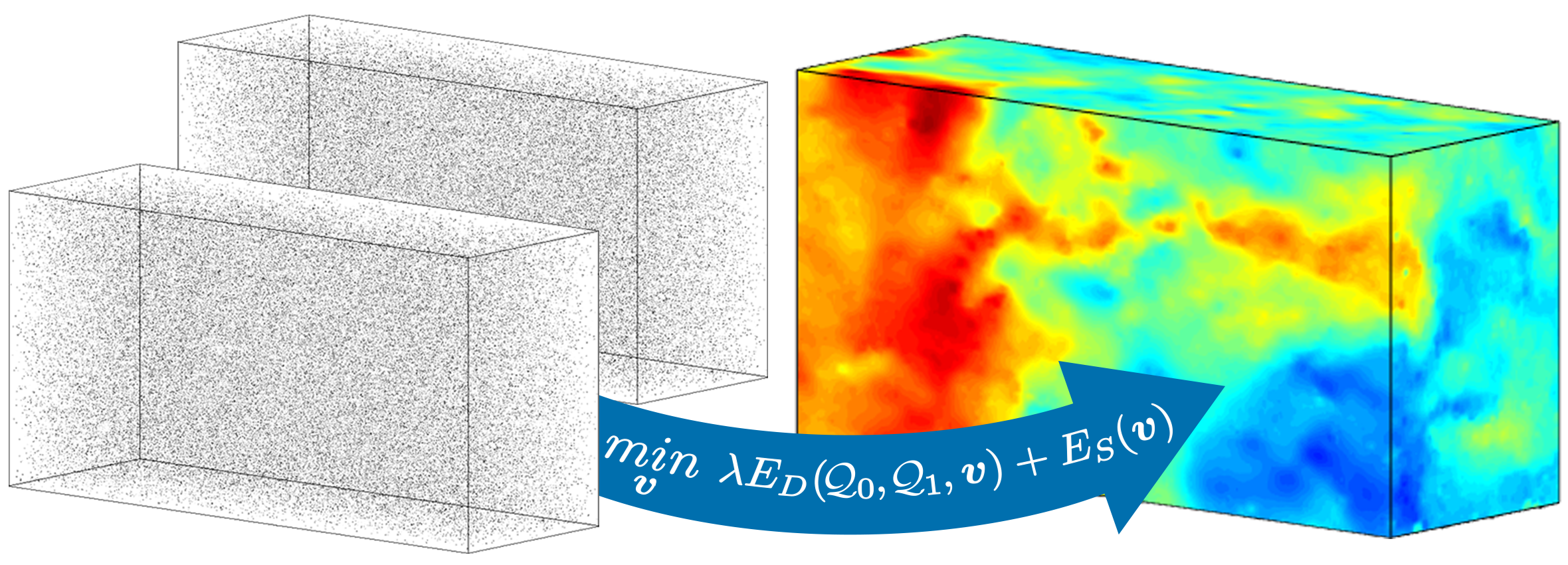}
 \caption{
Variational flow estimation from 3d particle reconstructions of two time steps.
 }
 \label{fig:teaser}
\end{figure}

\Fref{fig:teaser} depicts our basic setup, a 2-pulse variational
flow estimation approach from initial 3d reconstructions.

By operating densely in the 3D domain, our approach stands in stark
contrast with methods that build sparse particle trajectories over multiple
time-steps from image domain data~\cite{sch-16}.
Here, 3D information is used only for matching, in particular
to constrain the local search for matching particles in the temporal domain,
while regularization is introduced by demanding for smooth 2D trajectories.
Physical properties like incompressibility are much harder to model without the
consideration of 3D spatial proximity.
Hence, these methods have to fit a dense, volumetric representation to
the discovered, sparse particle tracks in a separate step.
There, relevant physical constraints~\cite{ges-16,schn-16} can be
included, yet cannot influence the particles' 3D motion any more.
While multi-frame analysis is certainly well-suited to improve the
accuracy and robustness of PIV over long observation times, we argue
that the basic case of two (or generally, few) frames is better
captured by dense flow estimation in 3D.

The proposed variational energy naturally fits well with modern
proximal optimization algorithms \cite{las-17}, and it produces only
little memory overhead, compared to conventional window-based tracking
in 3D \cite{els-06,Champagnat2011} without a-priori constraints.
In particular, we have shown in \cite{las-17} that the sparsity of the
particle data and the relative smoothness of the 3D motion field
requires highest resolution only for matching.  The 3D motion field
can therefore be parameterized at a lower resolution, such that the
whole optimization part requires about one order of magnitude less
memory than storing the intensity volume.

According to the classical variational formulation \cite{hor-81}, matching
should directly consider the smallest entities (pixel in 2D, voxel in 3D)
in the scene.
However, particles are sparsely distributed and of high ambiguity, such that
one rather compares a constellation of particles in a larger interrogation
volume, trading off robustness for matching accuracy.
Note that, compared to purely local methods \cite{els-06,Champagnat2011},
our global model also allows us to consider a smaller neighborhood
size for matching.

A popular alternative to matching raw spatial windows is to compress
the local intensity patterns into descriptors, \eg~\cite{lowe-04,dal-05}.
A descriptor can be designed to support robust matching of different entities,
like key-points in different views \cite{Mik-05}, pedestrians \cite{dal-05}
or to provide robustness against illumination changes \cite{Zab-94}.
Descriptors also improve optical flow computation in several
real-world settings \cite{liu-11}.  Here, we construct our own
descriptor specifically for 3D-PIV, that seeks to balance precision
and robustness.

Our descriptor can be constructed on-the-fly and does not require
storage of a high-resolution intensity volume, rather it works directly
on the locations and intensities of individual particles.
Recall that the resolution of the intensity voxel
grid~\cite{els-06,Champagnat2011,las-17}, is directly coupled to the
maximal precision of the estimated motion -- this dependency is
overcome by the new descriptor.
At typical particle densities, constructing descriptors on-the-fly
saves about three orders of magnitude in memory (compared to a
conventional voxel volume), but obtains comparable precision of the
motion field. Moreover, it allows one to fully exploit our
memory-efficient optimization procedure.

Although a particle-based representation can also be extracted from
tomographic methods~\cite{sch-11}, it appears more natural to directly
reconstruct the particles from the multi-view image data.
Iterative particle reconstruction methods (IPR), \eg~\cite{wie-13},
deliver exactly that, and often outperform Tomo-PIV methods especially in particle localization accuracy~\cite{wie-13}.
Building on these ideas, we pose IPR as an optimization problem. Since
such a direct energy formulation is highly non-convex, we add a proposal
generation step that instantiates new \emph{candidate} particles. That
step is alternated with energy optimization.
The energy is designed to explain intensity peaks that are consistent
across all views; while at the same time discarding spurious particles
that are not supported by sufficient evidence, with the help of a
sparsity prior.
Especially at large particle densities, our novel IPR-based technique
outperforms a tomographic baseline in terms of recall as well as
precision of particles (respectively number of missing and ghost particles), which in turn leads to significantly improved
fluid motion.

We also present a deep analysis of our IPR based particle
reconstruction model, where we analyze the effect of particle quality
and density on the motion estimation, and quantitatively study the
influence of the main parameters, including different layouts of the
sparse descriptor.

\section{Method}\label{sec:approach}
\begin{figure}[tb]
 \centering
 \includegraphics[width=\columnwidth]{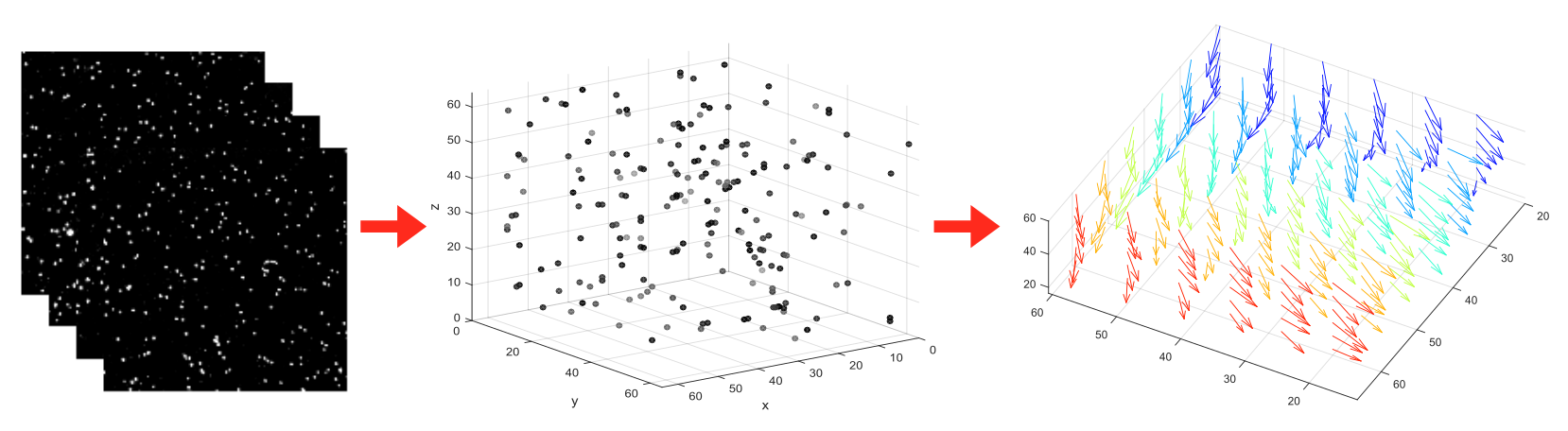}
 \caption{
Flow estimation pipeline. (left) 2D input data, (middle) 3D reconstruction, (right) 3D flow estimation from 2 time steps.
 }
 \label{fig:pipeline}
\end{figure}

We follow Horn and Schunk \cite{hor-81} and cast fluid motion
estimation directly as a global energy minimization problem over a
(cuboid) 3D domain $\Omega\subset\mathbb{R}^3$.
Emerging from variational principles, the energy of that inverse
problem can be divided into a data fidelity term and a regularizer.
As input, the basic model requires a set of 3D particles,
reconstructed individually for each time step,
see \fref{fig:pipeline}. It is described in more detail
in \Sec\ref{sec:3dreconstruction}.

As a baseline for particle reconstruction we utilize a well-known
tomographic reconstruction technique (MART) \cite{els-06}, which we
briefly describe next at the start of the technical section.
Afterwards, we present the novel, energy-based IPR solution, which
combines a simple image-based data term with an $\ell0$ sparsity. At
this point we also detail the optimization procedure, which
alternates between particle generation and energy minimization.
Then follows an explanation of the proposed sparse descriptor, which
exploits the low density of the particles within the fluid
in \Sec\ref{sec:sparseDescriptor}.

The second part, \Sec\ref{sec:volumetric}, describes the
volumetric model for fluid motion estimation.
We start with the regularizer which, following \cite{las-17}, is based
on the physical properties of the fluid, such as incompressibility and
viscosity.
Next we review the standard sum-of-squared-distances (SSD) data cost,
that we use as a baseline in our framework
in \Sec\ref{sec:volumetric}, and show how our sparse descriptor can
instead be integrated into the energy function as data cost.
Importantly, the energy is constructed to be convex, in spite of the
rather complex data and regularization terms. Hence, it is amenable
to a highly efficient optimization algorithm, which we describe at the
end of \Sec\ref{sec:volumetric}.

\subsection{3D Reconstruction}\label{sec:3dreconstruction}

The first objective of our pipeline is to reconstruct a set of
particles $\mathcal{Q}\!:=\!\{(p_l,c_l)\}_{l=1}^Q$, consisting of a set
of 3D positions $\mathcal{P}\!:=\!\{(p_l)\}_{l=1}^Q$,
$p_l\!\in\!\mathbb{R}^3$ and an associated set of intensities
$\mathcal{C}\!:=\!\{(c_l)\}_{l=1}^Q$, $c_l\!\in\!\mathbb{R}^+$, that
together define the data term of the energy function.
This is in contrast to voxel-based intensity volumes,
\eg \cite{els-06}. These dense volumetric representations can be obtained by
tomographic reconstruction methods, as traditionally used for
3D-PIV.
We compare two techniques to obtain the set of particles: a
tomographic method built on MART, and an iterative particle
reconstruction (IPR) \cite{wie-13} method that directly obtains sparse
3D particles and intensities from the 2D input images.
To extract 3D particle locations and intensities in continuous space
from the tomographic MART reconstruction, we perform peak detection
and subsequent sub-voxel refinement.
Note that the mapping from dense to sparse is invertible, in our
experiments we also utilize our IPR variant to generate a high
resolution intensity volume on which we define the voxel-based SSD
data term \cite{las-17}, as a baseline for our sparse descriptor.

In the following, we assume that the scene is observed by $k=1\ldots
K$ calibrated cameras in the images $I_k$ and define for each camera a
generic projection operator $\Pi_k$.
To keep the notation simple, we refrain from committing to a specific
camera model at this point.
However, we point out that fluid dynamics requires sophisticated
models to deal with refractions, or alternatively exhaustive
calibration to obtain an optical transfer
function \cite{wie-08,sch-12}.
Note that we reconstruct the particles individually for each time
step, and denote the two consecutive time steps as $t_0$ and $t_1$.

\paragraph{MART.}
Tomographic reconstruction assigns each voxel $i$ of a regular grid
$\mathcal{V}$ an intensity value $U(i)$.
A popular algebraic reconstruction technique, originally proposed
for three-dimensional electron microscopy, is MART \cite{GBH-70}.
MART solves the convex optimization problem:
\begin{equation}
\begin{aligned}
E_\textrm{\tiny MART}&(U):=
\sum\nolimits_{i\in\mathcal{V}} U(i) \log U(i) \textrm{ subject to }
\\
&\sum\nolimits_{i\in\mathcal{R}_j}\omega_{i,j} U(i)=I_k(j) \;\forall k\;\textrm{ and } U(i)\geq 0.
\end{aligned}
\end{equation}
Here, $I(j)$ denotes the observed intensity at pixel $j$,
$U(i)$ the unknown intensities of voxel $i$, and $\mathcal{R}_j$ is
the set of all voxels traversed by the viewing ray through pixel $j$.
The weight $\omega_{i,j}\!\in\![0,1]$ depends on the distance
between the voxel center and the line of sight and is essentially
equivalent to an uncertainty cone around the viewing ray, to account
for aliasing of discrete voxels.
For each pixel $j$ in every camera image $I_k$, and for each voxel
$i$, the following update step is performed, starting from
$U(i)^0:=1$:
\begin{equation}
U(i)^{n+1} = U(i)^{n}
 \cdot \Big(I_k(j) \Big/ \sum\nolimits_{i \in \mathcal{R}_p} \omega_{i,j} U(i)^{n}\Big)^{\omega_{i,j}}.
\end{equation}

After each MART iteration, we apply anisotropic Gaussian smoothing
($3\times 3\times 1$ voxels) of the reconstructed volume, to account
for elongated particle reconstructions along the $z$-axis due to the
camera setup~\cite{dis-13}.  Furthermore, we found it advantageous to
apply non-linear contrast stretching before the subsequent particle
extraction or flow computation:
$E_\mathrm{out} = (E_\mathrm{in})^\gamma$, with $\gamma=0.7$.

Given the intensity volume $U$, sub-voxel accurate sparse particle
locations $\mathcal{P}$ are extracted: first, intensity peaks are
found through non-maximum suppression with a $3\times3\times3$ kernel,
then we fit a quadratic polynomial to the intensity values in a
$3\times3\times3$ neighborhood.

\paragraph{IPR.}
As an alternative to memory-intensive tomographic approaches, IPR
methods iteratively grow a set of explicit 3D particles. They are
reported to achieve high precision and robustness, \cf\cite{wie-13}.
In contrast to the effective, but somewhat heuristic \cite{wie-13},
our variant of IPR minimizes a joint energy function over a set
of \emph{candidate} particles to select and refine the optimal
particle set. The energy minimization alternates with a candidate
generation step, where putative particles are instantiated by
analyzing the current residual images (the discrepancies between the
observed images and the back-projected 3D particle set).
Thus, in every iteration new candidate particles are proposed in
locations where the image evidence suggests the presence of
yet undiscovered particles.
Energy minimization then continually refines particle locations and
intensities to match the images, minimizing a data term
$E^D_\textrm{\tiny IPR}(\mathcal{P}, \!\mathcal{C})$. At the same
time, the minimization also discards spurious ghost particles, through
a sparsity prior $E^S_\textrm{\tiny IPR}$.
The overall energy is the sum of both terms, balanced by a parameter
$\eta$:
\begin{equation}\label{eq:EIPR}
E_\textrm{\tiny IPR}(\mathcal{P}, \mathcal{C}) :=
E^D_\textrm{\tiny IPR}(\mathcal{P}, \mathcal{C}) + \eta E^S_\textrm{\tiny IPR}(\mathcal{C})
\end{equation}

We model particles as Gaussian blobs of variance $\sigma$ and
integrate the quadratic difference between prediction and observed
image $I_k$ over the image plane $\Gamma_k$ of camera $k$:
\begin{equation}\label{eq:EIPR_D}
%
E^D_\textrm{\tiny IPR}(\mathcal{P}\!,\mathcal{C}) \!\!:=\!\! \sum_{k=1}^K \!\int_{\Gamma_k}\!\!\!\! \big| \mathcal{I}_k(x) \!-\!\! \sum_{l=1}^Q \!\Pi_k
( c_l\!\cdot\! \mathcal{N}(p_l,\sigma)(x) \! ) \big|_2^2 \dx.
\end{equation}
The model in \eref{eq:EIPR_D} assumes that particles do not possess
distinctive material or shape variations.  In contrast, we explicitly
model the particles' intensity, which depends on the distance to the
light source and changes only slowly.
Apart from providing some degree of discrimination between particles,
modeling the intensity allows one to impose sparsity constraints on the
solution, which reduces the ambiguity in the matching
process \cf \cite{Petra2009}.
In other words, we prefer to explain the image data with as few
particles as possible and set
\begin{equation}\label{eq:EIPR_S}
E^S_\textrm{\tiny IPR}(\mathcal{C}) := \sum_{l=1}^Q |c_l|_0 + \delta_{\{\geq 0\}} (c_l).
\end{equation}
The 0-norm in \eref{eq:EIPR_S} counts the number of particles with
non-zero intensity, and has the effect that particles falling below a
threshold (modulated by $\eta$) get discarded immediately.
In practical settings, the depth range of the observed volume is small
compared to the distance to the camera. Consequently, the projection
$\Pi$ can be assumed to be (almost) orthographic, such that particles
remain Gaussian blobs also in the projection.
Omitting constant terms, we can simplify the projection
in \eref{eq:EIPR_D} to
\begin{equation}\label{eq:ED_blob}
\Pi ( \mathcal{N}(\cdot,\sigma)(x) ) \!\approx\! \mathcal{N}(\Pi(\cdot),\sigma)(x)
\!\propto\! \frac{1}{\sigma^2}\!\exp
(\frac{-| \Pi(\cdot)\!-\!x |^2}{\sigma^2}),
\end{equation}
such that we are only required to project the particle center.
In practice, we can restrict the area of influence
of \eref{eq:ED_blob} to a radius of $3\sigma$.  While still covering
$99.7$\% of a particle's total intensity, this avoids having to
integrate particle blobs over the whole image in \eref{eq:EIPR}.

Starting from an empty set of particles, our particle generator operates
on the current residual images
\begin{equation}
I_{k,\operatorname{res}} \!:=\! \int_{\Gamma_k} I_k(x) \!-\! \sum_{l=1}^Q \!
\Pi_k(c_l\!\cdot\! \mathcal{N}(p_l,\sigma)(x)) \dx
\end{equation}
and proceeds as follows:
First, peaks are detected in all $k$ 2D camera images.  The detection
applies non-maximum suppression with a $3\times 3$ kernel and
subsequent sub-pixel refinement of peaks that exceed an intensity
threshold $\theta$.
For one of the cameras, $k\!=\!1$, we shoot rays through each peak and
compute their entry and exit points to the domain $\Omega$.
In the other views the reprojected entry and exit yield epipolar line
segments, which are scanned for nearby peaks.
For each constellation of peaks that can be triangulated with a
reprojection error $<\epsilon$ in all views, we generate a new
(candidate) particle.
The initial intensity is set depending on the observed intensity in
the reference view, and on the number of candidate particles to which
that same peak contributes.
In particular, for each of the $m$ detected particles
$\{p_l\}_{l=1}^m$ corresponding to the same peak in the reference, we set
$c_l\!:=\!\mathcal{I}_1(\Pi_1(p_l)) K / (K\!-\!1\!+\!m)$.

We run several iterations during which we alternate proposal
generation and energy optimization.
Over time, more and more of the evidence is explained, such that the
particle generator becomes less ambiguous and fewer candidates per
peak are found.
We start from a stricter triangulation error $\epsilon$ and increase the value in later iterations.

To minimize the non-convex and non-smooth energy in \eref{eq:EIPR}
we apply (inertial) PALM \cite{Bolte2014,poc-16}.
Being a semi-algebraic function \cite{Bolte2014,Attouch2013}, our energy
\eref{eq:EIPR} fulfills the Kurdyka-Lojasiewicz property \cite{Bolte2006},
so the sequence generated by PALM is guaranteed to converge to a critical
point of \eref{eq:EIPR}.
To apply PALM we arrange the variables in two blocks, one for the
locations of the particles $\mathcal{Q}$ and one for their
intensities. This gives two vectors
$\mathbf{p}\!:=\!(p_1\trans,\ldots,p_Q\trans)\trans\!\in\!\mathbb{R}^{3Q}$
and
$\mathbf{c}\!:=\!(c_1,\ldots,c_Q)\trans\!\in\!\mathbb{R}^{Q}$.
The energy is already split into a smooth part \eref{eq:EIPR_D} and
non-smooth part \eref{eq:EIPR_S}, where the latter is only a function
of one block ($\mathbf{c}$).
In this way, we can directly apply the block-coordinate descent scheme
of PALM.
The algorithm iterates between the two blocks $\mathbf{p}$ and
$\mathbf{c}$ and executes the steps of a proximal
forward-backward scheme.
At first, we take an explicit step \wrt one block of variables
$z\!\in\!\{\mathbf{p},\mathbf{c}\}$ on the smooth part
$E^D_\textrm{\tiny IPR}$, then take a backward (proximal) step on the
non-smooth part $E^S_\textrm{\tiny IPR}$ \wrt the same block of
variables -- if possible.

That is, at iteration $n$ and $z \in \{ \mathbf{p}, \mathbf{c} \}$, we
perform updates of the form
\begin{equation}\label{eq:PFB_step}
\begin{aligned}
z^{n+1} = &\textrm{prox}^{ E^S_\textrm{\tiny IPR}}_t(z) := \argmin_y  E^S_\textrm{\tiny IPR}(y) + \frac{t}{2} \| y-z \|^2
\\
& \textrm{ with } \quad z = z^n - t^{-1} \nabla_z  E^D_\textrm{\tiny IPR}(\cdot,z^n),
\end{aligned}
\end{equation}
with a suitable step size $1/t$ for each block of variables.

Throughout the iterations, we further have to ensure that the step
size in the update of a variable block
$z\!\in\!\{\mathbf{p},\mathbf{c}\}$ is chosen in accordance with the
Lipschitz constant $L_z$ of the partial gradient of function
$E^D_\textrm{\tiny IPR}$ at the current solution:
\begin{equation}\label{eq:Lipshitz_condition} 
\|\nabla_z E^D_\textrm{\tiny IPR}\!(\!\cdot,\!z_1) \!-\! \nabla_z E^D_\textrm{\tiny IPR}(\!\cdot,\!z_2)\| \!\leq\! L_z(\!\cdot,\!\cdot) \|z_1\!-\!z_2\| \,\forall z_1,\!z_2.
\end{equation}
In other words, we need to verify whether the step size $t$ in
\eref{eq:PFB_step}, used for computing the update $z^{n+1}$,
fulfills the descent lemma \cite{Bertsekas1989}:
\begin{equation}\label{eq:Lipshitz_test}
\begin{aligned}
E_\textrm{\tiny IPR}( \cdot, z^{n+1} ) \leq& E_\textrm{\tiny IPR}( \cdot, z^n ) + \langle \nabla_z E^D_\textrm{\tiny IPR}(\cdot, z^n ), \\ & z^{n+1}\!-\!z^{n}\rangle
 + \frac{t}{2} \|z^{n+1}-z^n\|^2.
\end{aligned}
\end{equation}
Otherwise the update step has to be recalculated with a smaller step size.
In practice, the Lipschitz continuity of the gradient
\eref{eq:Lipshitz_test} of $E^D_\textrm{\tiny IPR}$ only has to be
checked locally at the current solution.
Hence, we employ the back-tracking approach of \cite{Beck2009},
which is provided in pseudo-code in \Alg\ref{alg:ipalm}.
\setlength{\textfloatsep}{3.pt}
\begin{algorithm}[tb]
\caption{iPalm implementation for energy (\ref{eq:EIPR})}\label{alg:ipalm}
\begin{algorithmic}[1]
\Procedure{ipalm}{$\mathbf{p}^0,\mathbf{c}^0}$ 
\State
$
\mathbf{p}^{-1} \gets \mathbf{p}^{0};
\mathbf{c}^{-1} \gets \mathbf{c}^{0};$
\State
$\tau\gets\frac{1}{\sqrt{2}};
L_\mathbf{p}\gets 1;L_\mathbf{c}\gets 1;$ 
\For{n:=0 to $n_\textrm{steps}$}
\State $\mathbf{\hat{p}} \gets \mathbf{p}^n + \tau (\mathbf{p}^n-\mathbf{p}^{n-1}); $ // inertial step
\While{true}
\State
$
\mathbf{p}^{n+1} := \mathbf{\hat{p}} - 1/L_\mathbf{p} \nabla_\mathbf{p} E^D_\textrm{\tiny IPR}(\mathbf{\hat{p}}, \mathbf{c}^n);
$
\If {$\mathbf{p}^{n+1}, L_\mathbf{p}$ fulfill \eref{eq:Lipshitz_test}}
\textbf{break};
\Else{ $L_\mathbf{p}=2 L_\mathbf{p}$};
\EndIf
\EndWhile
\State $\mathbf{\hat{c}} \gets \mathbf{c}^n + \tau (\mathbf{c}^n-\mathbf{c}^{n-1}) $; // inertial step
\While{true}
\State
$
\mathbf{c} := \mathbf{\hat{c}} - 1/L_\mathbf{c} \nabla_\mathbf{c} E^D_\textrm{\tiny IPR}(\mathbf{p}^{n+1}, \mathbf{\hat{c}});
$
\State
$
\mathbf{c}^{n+1}:= \textrm{prox}^{F_\mathbf{c}}_{L_\mathbf{c}}(\mathbf{c})$; // \eref{eq:prox_c}
\If {$\mathbf{c}^{n+1}, L_\mathbf{c}$ fulfill \eref{eq:Lipshitz_test}}
\textbf{break};
\Else{ $L_\mathbf{c}=2 L_\mathbf{c}$};
\EndIf
\EndWhile
\EndFor
\EndProcedure
\end{algorithmic}
\afterpage{\global\setlength{\textfloatsep}{8pt}}
\end{algorithm}

For simple proximal operators, like ours, it is recommended to also
increase the step sizes as long as they still
fulfill \eref{eq:Lipshitz_test}, instead of only reducing them when
necessary (lines 9,15 in \Alg\ref{alg:ipalm}). Taking larger steps per
iteration speeds up convergence.
We also take inertial steps \cite{poc-16} in \Alg\ref{alg:ipalm} (lines 5,10).
Inertial methods can provably accelerate the convergence of
first-order algorithms for strongly convex energies,
improving the worst-case convergence rate from
$O(1/n)$ to $O(1/n^2)$ in $n$ iterations, \cf \cite{Beck2009}.
Although there are no such guarantees in our non-convex case, these steps
significantly reduce the number of iterations in our algorithm by a factor
of five, while leaving the computational cost per step practically untouched.

The proximal step on the intensities $\mathbf{c}$ can be solved
point-wise. It  can be written as the 1D-problem
\begin{equation}
\begin{aligned}
\textrm{prox}^{E^S_\textrm{\tiny IPR}}_t (\bar{c}) := 
\argmin_c \eta |c|_0 + \delta_{\{\geq0\}} (c) + \frac{t}{2} | c-\bar{c} |^2,
\end{aligned}
\end{equation}
which admits for a closed-form solution:
\begin{equation}\label{eq:prox_c}
\textrm{prox}^{E^S_\textrm{\tiny IPR}}_t (\bar{c}) :=
\left\{\!\!
   \begin{array}{c@{\hspace{0.2cm}}l}
     0          & \textrm{if } t \bar{c}^2 \!\!<\! 2\eta \textrm{ or } \bar{c}\!<\!0\\
     \bar{c}    & \textrm{else},
   \end{array}
   \right. .
\end{equation}

\subsection{Sparse Descriptor}\label{sec:sparseDescriptor}

In PIV, particle constellations are matched between two time steps.
To that end, an interrogation window at the location for which one
wants to determine velocity is considered as the local ``descriptor''.
Velocity information is obtained from the spatial displacement between
corresponding patches, typically found via \emph{(normalized) cross
correlation} ((N)CC).
For 3D-PIV this requires the storage of full intensity voxel
volumes, like those obtained with tomographic reconstruction.
Discretizing the particle information at this early stage has the
disadvantage that the resolution of the voxel grid directly limits the
maximal precision of the motion estimate.
Especially when considering multiple time steps, storing intensity
voxel volumes at high resolution is demanding on memory, although the
actual particle data is sparse.
Moreover, large window sizes are needed to obtain robust
correspondence, which (over-)smoothes the output, trading off
precision for robustness.
Notably, in practical 3D-PIV these required interrogation volumes are
very large, \eg Elsinga \etal \cite{els-06} recommend $41^3$.
For typical particle densities, such an interrogation window
contains only 25 particles on average.
Finally, IPR-type methods reportedly outperform MART in terms of
reconstruction quality, and directly deliver sparse particle output.

These findings motivate us to investigate a new feature descriptor
that directly works on the sparse particle data.
In order to replace the conventional correlation window, our novel
descriptor must be defined at arbitrary positions in $\Omega$ and
lead to sub-voxel accurate measurements.
To use the descriptor in our variational setup we further demand that
it be differentiable.
Finally, the descriptor should be independent of the distance metric,
such that we can use standard measures like SSD or (negative) NCC.

Similar to an interrogation window, the descriptor considers all
particles within a specified distance around its center location.
Its structure is defined by a radius $r$, the arrangement of a set of
vertices $H$, and an integer count $k$.
All particles within a distance smaller than $r$ contribute to
the descriptor vector.
For such a particle $q=(p,c)$, we further define the set of $k$
nearest neighbors in $H$, $\mathcal{N}^k_H(p)$, to which 
the intensity
$c$ of the particle is distributed.
We weight the contribution of $q$ to the vertex $h\in\mathcal{N}^k_H(p)$
with a radial distance function $exp(-\|p-h\|/2)$
and normalize the $k$ weights induced by the particle to sum to one.
In computer graphics, such a 3D-to-2D projection by weighted
soft-assignment is called \emph{splatting}.
Apart from higher precision of the descriptor,
\cf \cite{lowe-04,dal-05}, the splatting strategy leads to smoother
descriptor changes when the nearest neighbor set changes after a
spatial displacement of the particle. More importantly, it guarantees
that the descriptor is at least piecewise differentiable.

\begin{figure}[tb]
 \centering
 \includegraphics[width=0.48\columnwidth]{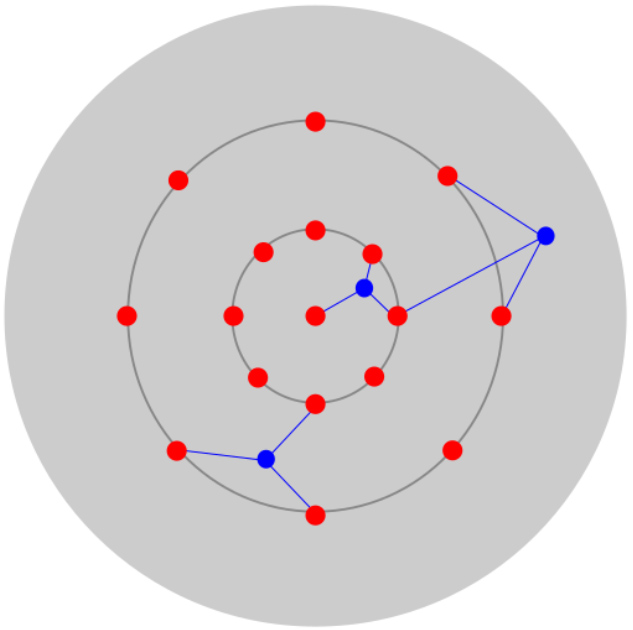}
 \includegraphics[width=0.48\columnwidth]{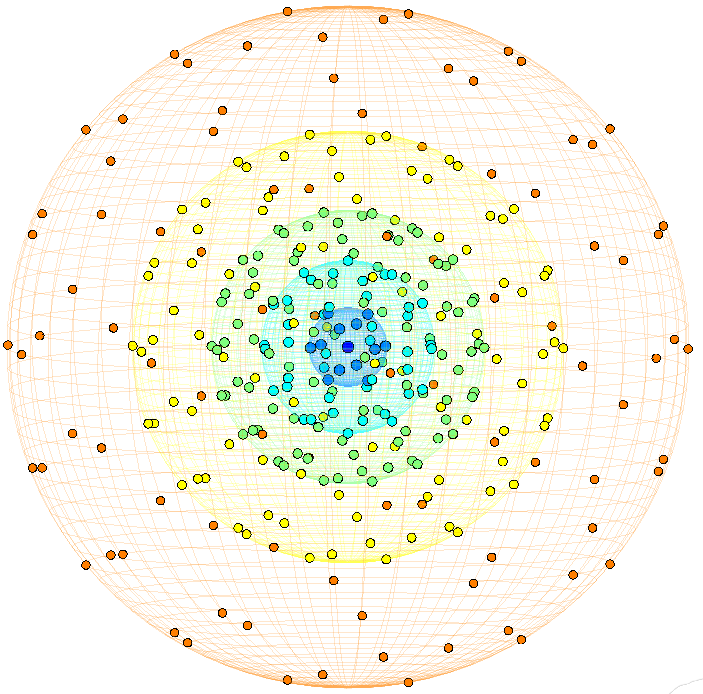}
 \caption{
 Left: Schematic visualization of a 2D sparse descriptor with descriptor
 vertices (red) and particle locations (blue),
 whose intensities are splatted over the 3 nearest neighbors.
 Right: 3D descriptor vertices (with layer based color-code).
 }
 \label{fig:descriptor}
\end{figure}

Note, when arranging $H$ in a regular orthogonal grid and setting
$k\!=\!8$, the descriptor closely resembles a conventional
interrogation volume.
Instead, we propose to arrange the vertices on spherical shells with
different radii. A 2-dimensional illustration with $k=3$ is shown in
\fref{fig:descriptor}. Vertices are depicted in red and particles in blue.

Any descriptor is constructed to fulfill some set of desirable
properties~\cite{fro-04,sco-07,yen-14}.  Designing one specifically
for our application has the advantage that it can be tuned to the
specific scenario.
In our case we do not seek rotational invariance, and all directions
should be considered equally (i.e.~same vertex density on poles and
equator).  Further, our input is unstructured (in contrast to, for
instance, laser scan points that adhere to 2D surfaces).
Similar to~\cite{yen-14} we prefer to distribute the vertices $H$
uniformly on the surface of multiple spheres. Accordingly, we place
them at the corners of an icosahedron (or subdivisions of it) for each
sphere.
We further seek equidistant spacing between neighboring spheres and chose
the distance of the radii in accordance with the distances of vertices
on the inner shell.
Because particles close to the center are more reliable for its
precise localization, we place more points towards the descriptor
center, for a more fine-grained binning of close particles.
In contrast, particles that are further away are more likely have
different velocity and are splatted over a wider area.  Hence, these
particles mainly contribute to disambiguate the descriptor.
The precision argument also suggests to weight particle closer to the
center more strongly than those further away.  Similarly, we
compensate the fact that the vertices near the center represent a
smaller volume than those further away, by weighting their
contributions accordingly.
To evaluate our descriptor centered at a given location in $\Omega$,
we must detect the particles within a radius $r$ of that location,
and also find the $k$ nearest neighbours for each particle within $r$.
Recall that in our scheme we pre-compute a set of particles $\mathcal{Q}$
consisting of location and intensity at both time steps.
To find the particles we construct a
KD-tree~\cite{fri-77} once for each time step, to accelerate the search.
In our low-dimensional search domain ($\Omega\subset\mathbb{R}^3$),
with limited number of particles ($\approx 3$ orders of magnitude
smaller than voxels in an equivalently accurate intensity volume), the
complexity of the search is quite low, $O(\log |\mathcal{Q}|)$.
To find the $k$ nearest vertices for each contributing particle we can
use a mixed strategy:
We partition the descriptor volume into a voxel grid and pre-compute
the nearest neighbors for the corners of the voxels, storing the $k$
nearest neighbor if that set is unique. In this way, the $k$-NN search
can be done in constant time for a large majority of all cases.
To handle the rare case of particles falling in voxels without a
unique neighbor set, we pre-compute a KD-tree for the rather small set
of vertices $H$ and query the neighbors of the concerned particles.
We use the KD-tree implementation proposed in \cite{muj-14}.
Overall, we find that, in spite of the need to frequently query
nearest neighbours, our descriptor can be evaluated efficiently for
any realistic particle density, while saving a lot of memory.

We have tested different layouts and have decided for a structure of
six layers.  The descriptor vertices are depicted
in \fref{fig:descriptor} on the right.  On each layer, points are
positioned on an icosahedron or subdivisions of it. The point and
layer arrangement is described in \tref{tab:descriptorStructure}.  In
total the descriptor is a vector $D_\mathcal{Q}(x)\in \mathbb{R}^{331}$ with
331 dimensions (vertices).
We empirically chose a radius of $r=10.5$. Note that this leads to a larger influence area than the interrogation volume of $13^3$ suggested in  \cite{las-17}. However, in \cite{las-17} a squared area is used for matching whereas our descriptor has a spherical form. Furthermore, as stated above, particles closer to the descriptor center are weighted higher than further away particles.

Finally we splat the intensity value of each particle over $k=5$ neighbors
in the descriptor.
%
Vertices are weighted according to their distance to the descriptor
center by $w_g=exp(-\|g\|/10)$ and subsequent normalization such
that $\sum w_g\!=\!1$.
We point out that an interrogation volume of comparable metric size,
i.e.~$17^3$ voxels, would consist of 4913 voxels to represent the same
particle distribution captured in our 331-dimensional descriptor.

\begin{table}
\caption{\label{tab:descriptorStructure}Layered structure of our descriptor.}
\setlength\tabcolsep{0.16cm}
\begin{tabular}{@{}ccccc}
\hline
\rowcolor{gray!10}
& & & Distance to & Distance to \\
\rowcolor{gray!10}
Layer & \#Points & Radius & neighbor points & lower layer\\
\hline
0 & 1 & 0 & - & - \\
1 & 12 & 1 & 1.052 & 1 \\
2 & 42 & 2.205 & 1.205 & 1.205  \\
3 & 92 & 3.484 & 1.278  & 1.278 \\
4 & 92 & 5.503 & 2.019 & 2.019 \\
5 & 92 & 8.693 & 3.190 & 3.190 \\
\hline
\end{tabular}
\end{table}

\subsection{Volumetric Flow}\label{sec:volumetric}

Motion estimation starts from two particle sets
$\mathcal{Q}_0,\mathcal{Q}_1$ for consecutive time steps $t^0$ and
$t^1$, with particles located in the cuboid domain
$\Omega\subset\mathbb{R}^3$.
Our goal is to reconstruct the 3D motion field
$\mathbf{v}:\Omega\rightarrow\mathbb{R}^3$, represented as a
functional $\mathbf{v}=(u,v,w)\trans$, which maps points $p$ inside
the domain $\Omega$ at $t^0$ to points $(p+\mathbf{v})$ in
$\mathbb{R}^3$ (but not necessarily inside $\Omega$) at $t^1$.
Just like 2D optical flow, the problem is severely ill-posed, if the
estimation were to be based solely on the data
term
$E_D(\mathcal{Q}_0,\mathcal{Q}_1,\mathbf{v})$.
As usual in optical flow estimation, we thus incorporate prior
information into the model, in the form of a regularizer
$E_S(\mathbf{v})$.  The optimal flow field can then be found by
solving the energy minimization problem
\begin{equation}\label{eq:energy}
\operatorname*{arg\,min}_{\mathbf{v}}
\lambda E_D(\mathcal{Q}_0,\mathcal{Q}_1,\mathbf{v}) + E_S(\mathbf{v}) 
\end{equation}
With the help of a convex approximation of the (non-convex) data term,
\eg, a first-order or convexified second-order Taylor approximation,
we can cast \eref{eq:energy} as a saddle-point problem and minimize it
with efficient primal-dual algorithms such as~\cite{cha-11}.
In spite of the high frame-rate of PIV cameras, particles travel large
distances between successive frames.
We use the standard technique to handle these large motions and embed
the computation in a coarse-to-fine scheme \cite{bro-04}, starting
from an approximation of $E_D$ at lower spatial resolution and
gradually refining the motion field.
Downsampling the spatial resolution smoothes out high-resolution
detail and increases the radius over which the linear approximation is
valid.

We employ an Eulerian representation for our functional $\mathbf{v}$,
although our sparse descriptor would equally allow for a Lagrangian
representation, for instance employing
\emph{Smoothed Particle Hydrodynamics} \cite{Monaghan2005,Adams2007}.
Naturally, we decide to evaluate the data term on the same regular
grid structure, and the proposed descriptor
(\sref{sec:sparseDescriptor}) is used as a drop-in replacement for the
interrogation volume in the classical Tomo-PIV setup.

\paragraph{Data Term.}
Recall that we can apply any distance measures $S$ on top of our
descriptor $D_\mathcal{Q}(x)$.
Here, we have tested \emph{sum of squared differences}
$S:=\operatorname*{SSD}$ and negative \emph{normalized
cross-correlation} $S:=-\operatorname*{NCC}$.
In previous evaluations for 3D-PIV \cite{las-17}, more robust
cost functions that are popular in the optical flow
literature \cite{vog-13} performed worse than both of the measures
above, presumably due to the controlled contrast and lighting.
Our (generic) similarity function
$S:\mathbb{R}^{331}\times\mathbb{R}^{331}\to\mathbb{R}$ takes
two descriptor vectors $D_\mathcal{Q}\in\mathbb{R}^{331}$ as input
and outputs a similarity score,
which is integrated over the domain to define the data cost:
\begin{equation}\label{eq:dataSparse}
E_D^S(\mathcal{Q}_0,\mathcal{Q}_1,\mathbf{v}):=
\int_\Omega S (D_\mathcal{Q}^{t_0}(x), D_\mathcal{Q}^{t_1}(x))\dx
\end{equation}

As a baseline we further define the $\operatorname*{SSD}$ data cost
$E_{U,D}^{\operatorname*{SSD}}$, evaluated within an interrogation
window $\mathcal{N}$ in a high-resolution intensity voxel
space (in our experiments, generated by MART).
In \cite{las-17} we showed that for high-accuracy motion estimates
it is crucial to use a finer discretization for the data term than for
the discretized Eulerian flow field $\mathbf{v}$.
In that case the corresponding data cost $E_{U,D}^{\operatorname*{SSD}}$
takes the form
\begin{equation}\label{eq:dataDense}
\begin{aligned}
E_{U, D}^{\operatorname*{SSD}}&(U_0,U_1,\mathbf{v}) = \\
\int_\Omega &\int_\Omega [U_0(x) - U_1(x + \mathbf{v}(z))] ^2 B_\mathcal{N}(z-x) \mathrm{d}x\mathrm{d}z,
\end{aligned}
\end{equation}
where $B_\mathcal{N}$ is a box filter of width $|\mathcal{N}|$ and
$U_0,U_1:\Omega\rightarrow\mathbb{R}^+$ are intensity volumes, defined over
the same cuboid domain $\Omega$ but sampled at a different discretization level.

\paragraph{Regularization.}
In the variational flow literature \cite{hor-81,bro-04,wer-10}
regularization usually amounts to simply smoothing the flow field.
The original smoothness term from~\cite{hor-81} is quadratic
regularization (QR), later the more robust Total Variation
(TV)~\cite{rud-92,zac-07} was often preferred.
An investigation in \cite{las-17}) revealed, however,
that TV does not perform as well for 3D fluid motion.
Hence, we only provide a formal definition of the QR regularizer:
\begin{equation}\label{eq:regularizers}
\operatorname{QR}(\mathbf{v}) = \frac{1}{2} \int_{\Omega} |\nabla u|^2 + |\nabla v|^2+ |\nabla w|^2 \mathrm{d}x. 
\end{equation}
Here we choose $|\cdot|:= \|\cdot\|_2$.

In this work, we are mainly concerned with incompressible flow,
which is divergence-free.
Prohibiting or penalizing divergence is thus a natural way to
regularize the motion field.

Let us recall that the motion field approximates the displacement of
particles over a time interval $\Delta t$, reciprocal to the camera frame rate.
\Ie, $\mathbf{v}\approx\Delta t \mathbf{\bar{v}}$,
where $\mathbf{\bar{v}}$ denotes the velocity field. 
We now investigate the stationary Stokes equations:
\begin{equation}\label{eq:StatStokes}
-\mu\Laplace\mathbf{\bar{v}} + \nabla p = \mathbf{f} \quad \textrm{subject to } \div \mathbf{\bar{v}} = 0,
\end{equation}
where the generalized Laplace operator $\Laplace$ operates on each
component of the velocity field $\mathbf{\bar{v}}$ individually.
The equation \eref{eq:StatStokes} describes an external force field
$\mathbf{f}$ that leads to a deformation of the fluid.
The properties of the fluid, incompressibility and viscosity
(viscosity constant $\mu$), prevent it from simply following $\mathbf{f}$.
To arrive at an equilibrium of forces, the pressure field $p$ has to
absorb differences in force and velocity field $\mathbf{\bar{v}}$.
Compared to the full Navier-Stokes model of fluid dynamics,
\eref{eq:StatStokes} lacks the transport equations and, hence,
does not consider inertia.
Nevertheless, for our basic flow model with only 2 frames
\eref{eq:StatStokes} provides a reasonable approximation, because of the short
time interval.

We can identify the stationary Stokes equations \eref{eq:StatStokes}
as the Euler-Lagrange equations of the following energy:
\begin{equation}\label{eq:regularizerMotiv}
\begin{aligned}
\operatorname*{min}_{\mathbf{\bar{v}}}\ \operatorname*{max}_{p}
\int_\Omega &
\frac{\mu}{2}
(|\nabla \bar{u}|^2 + |\nabla \bar{v}|^2+ |\nabla \bar{w}|^2)
+ \\ &
\langle \div\mathbf{\bar{v}}, p \rangle - \langle \mathbf{\bar{v}}, \mathbf{f} \rangle \mathrm{d}x
\end{aligned}
\end{equation}

To come back to our time-discretized problem, and to justify the construction of
a suitable regularizer for our motion field $\mathbf{v}$, we chose the time unit as
$\Delta t=1$, without loss of generality. 
In the saddle-point problem above, the role of the Lagrange multiplier for
the incompressibility condition is taken by the pressure $p$.
In our reconstruction task, we do not know the force field, but we do
observe its effects, such that the role of the term
$\langle \mathbf{v},f \rangle$ is filled by the data term.  Hence, we
can interpret the remaining terms as a regularizer.
In particular, \Eq\eref{eq:regularizerMotiv} suggests to utilize
a quadratic regularizer on the flow field, and to employ the
incompressibility condition as a hard constraint:
\begin{equation}\label{eq:regularizers_QRDinfty}
\begin{aligned}
\operatorname{QRD_\infty}(\mathbf{v}) = \frac{1}{2} \int_{\Omega} |\nabla u|^2 + |\nabla v|^2+ |\nabla w|^2
+ \delta_{\{0\}} (\div\mathbf{v}) \mathrm{d}x. 
\end{aligned}
\end{equation}
Here, we let $\delta_C$ denote the indicator function of the convex
set $C$.
This physically motivated regularization scheme is compatible with our
optimization framework, and leads to the best results for our data.
Still, without the hard constraint the
regularizer \eref{eq:regularizerMotiv} would be strongly convex,
making it possible to accelerate the optimization~\cite{cha-11},
similar to \Alg\ref{alg:ipalm}.
Therefore, we also consider a version where we keep the quadratic
regularization of the gradients, but replace the incompressibility
constraint with a soft penalty.
\begin{equation}\label{eq:regularizers_QRDA}
\operatorname{QRD_\alpha}(\mathbf{v}) \!=\! \frac{1}{2} \int_{\Omega}\! |\nabla u|^2 + |\nabla v|^2+ |\nabla w|^2 +\alpha (\div\mathbf{v})^2 \mathrm{d}x.
\end{equation}
Larger $\alpha$ lead to similar regularization effects as
\eref{eq:regularizerMotiv}, but also require more iterations
to reach convergence.

\paragraph{Estimation Algorithm.}
To discretize the energy functional \eref{eq:energy} we partition the
domain $\Omega$ into a regular voxel grid $\mathcal{V}:=\{1\ldots
N\} \times \{1\ldots M\}\times \{1\ldots L\}$.
Our objective is to assign a displacement vector
$\mathbf{v}_i:=\mathbf{v}(i)\in\mathbb{R}^3$ to each voxel
$i\in\mathcal{V}$, so $\mathbf{v}\in V,\; V:=\mathbb{R}^{3NML}$.
We briefly review the primal-dual approach~\cite{cha-11}
for problems of the form
\begin{equation}\label{eq:primal}
  \operatorname*{min}_{\mathbf{v}}\; F(D \mathbf{v}) + G(\mathbf{v})\;.
\end{equation}
Here $D\;:\; V \to Y$ is a linear operator whose definition depends on
the form of the regularizer.
$G$ denotes the discretized version of the data term from \eref{eq:dataSparse}
or \eref{eq:dataDense}, and $F$ one of the investigated regularizers from
(\ref{eq:regularizers},~\ref{eq:regularizers_QRDinfty},~\ref{eq:regularizers_QRDA}).
In case of QR, the linear mapping $D$ 
approximates the spatial gradient of the flow in each coordinate direction
via finite differences, which leads to
$Y:=Y^1\times Y^2\times
Y^3=\mathbb{R}^{3NML}\times\mathbb{R}^{3NML}\times\mathbb{R}^{3NML}$.
If we base our regularizer also on the divergence of the flow field,
either as hard \eref{eq:regularizers_QRDinfty} or
as soft \eref{eq:regularizers_QRDA} constraint,
we expand $D$ with a linear operator for the 3D divergence that is
based on finite (backward) differences.
We arrive at $Y:=Y^1\times Y^2\times Y^3\times Y^4$.
In that case $Y^4\in \mathbb{R}^{NML}$ contains the dual variables for
the incompressibility constraint, respectively penalty; the pressure
field \eref{eq:StatStokes}.
For convex $F$ we can convert \eref{eq:primal} to the saddle-point form
%
\begin{equation}\label{eq:optim}
\operatorname*{min}_{\mathbf{v} \in V} \ \operatorname*{max}_{\mathbf{y} \in Y} \;\langle D\mathbf{v},\mathbf{y} \rangle - F^*(\mathbf{y}) + G(\mathbf{v})\;,
\end{equation}
where we use
$F^*(\mathbf{y}):=\sup_{\mathbf{z}\in Y} \mathbf{z}\trans \mathbf{y} - F(\mathbf{z})$
to denote the convex conjugate of $F$.
In this form the problem can be solved by
iterating the steps of a proximal forward-backward scheme,
alternating between primal $\mathbf{v}$ and dual variables $\mathbf{y}$:
\begin{equation}\label{eq:pd_iterates}
\begin{aligned}
\mathbf{v}^{n+1} =& (I +  \tau \partial G)^{-1} ( \mathbf{v}^n- \tau D\trans \mathbf{y}^n )
\\
\mathbf{y}^{n+1} =& (I +  \sigma \partial F^*)^{-1} ( \mathbf{y}^n + \sigma D (2\mathbf{v}^{n+1}-\mathbf{v}^{n} )).
\end{aligned}
\end{equation}
The step sizes $\tau$ and $\sigma$ are set such that $\tau\sigma\leq 1/\|D\|^2_2$.
In the saddle-point form, data and smoothness terms are decoupled and
the updates of the primal and dual variables can be solved point-wise,
\cf~\cite{cha-11}.

The proximal operator of $F^*$ for
$\operatorname{QR}$ and $\operatorname{QRD_\alpha}$ is a pixel-wise
operation that can be applied in parallel  $\forall i\,\in\,\mathcal{V}$:
\begin{equation}
\Big( (I + \sigma \partial F^*)^{-1} ( \mathbf{y}^k ) \Big)_i = \mathbf{y}^k_i
\;\frac{\alpha^k}{\alpha^k+\sigma},
\end{equation}
with $\alpha^k=1, \forall k$, except in the case of $\operatorname{QRD_\alpha}$
where we additionally have $\alpha^4:=\alpha$.
If we employ hard incompressibility constraints, elements from
$Y^4$
remain unchanged and are solely affected by the explicit gradient steps.
For further details refer to~\cite{cha-11,vog-13}.

To discretize our data terms for the sparse descriptor, we use a simple
first order approximation, fitting a linear function
in the vicinity of the current solution.
This leads to the discretized data cost, expanded at the current solution
for the motion vector at voxel i
$\mathbf{v}_{0,i}:=\mathbf{v}(i)$ for
$S\in\{\lambda\operatorname*{SSD},-\lambda\operatorname*{NCC}\}$:
\begin{equation}\label{eq:disc_desc_data}
\begin{aligned}
S &(D_\mathcal{Q}^{t_0}(i), D_\mathcal{Q}^{t_1}(i+\mathbf{v}_i)) \approx
G_{\operatorname*{S},\mathcal{Q}_0,\mathcal{Q}_1}(i,\mathbf{v}_i) := \\ &
S (D_\mathcal{Q}^{t_0}(i), D_\mathcal{Q}^{t_1}(i+\mathbf{v}_{0,i})) +
\langle \mathbf{f}_i, \mathbf{v}_{0,i}-\mathbf{v}_{i}\rangle
\end{aligned}
\end{equation}
This generic treatment allows for a simple implementation and evaluation of
different distance metrics. For instance, the negative NCC cost function is
not convex and does not possess a closed form solution for the proximal map.
Although also a (more accurate) second-order Taylor expansion could be
used to build a convex approximation~\cite{wer-10,vog-13}, we opt for a
numerical solution.
Due to the rather low evaluation time of our descriptor, the
additional computation time for a numerical solution is quite small.
In our implementation, we approximate the linear function
$\langle\mathbf{f}_i,\cdot\rangle,  \mathbf{f}_i\in\mathbb{R}^3$
via least-squares fit to the six values
$S (D_\mathcal{Q}^{t_0}(i), D_\mathcal{Q}^{t_1}(i+\mathbf{v}_{0,i}\pm h e_k))$,
with $e_k$ being the unit vector in direction $k$ and a small $h$.
To find a solution for the proximal step one must solve the following
quadratic problem per voxel $i$:
\begin{equation}
\begin{aligned}
 \left((I +  \tau \partial G)^{-1}(\hat{\mathbf{v}})\right)_i = &
 \operatorname*{arg\,min}_{\mathbf{v}_i}
\frac{1}{2\tau} (\hat{\mathbf{v}}_i - \mathbf{v}_i)^2
+ \\&
G_{\operatorname*{S},\mathcal{Q}_0,\mathcal{Q}_1}(i,\mathbf{v}_i).
\end{aligned}
\end{equation}

We also briefly review the proximal step for the SSD data term
\eref{eq:dataDense}, operating on the intensity voxel volumes $U_0,U_1$.
Discretizing \eref{eq:dataDense} leads to the following cost for a voxel $i$:
\begin{equation}\label{eq:ssd}
\operatorname*{SSD}(i,\mathbf{v}_i) =\!\! \sum_{j\,\in\,\mathcal{N}(i)}\!
|U_0(j)-U_1(j+\mathbf{v}_i)|^2 \omega(j-i).
\end{equation}
with $\omega(j-i) \!=\! \frac{1}{|\mathcal{N}(i)|}$.
A direct convex approximation would be to use first-order
Taylor-expansion around $j+\mathbf{v}_i$ on all the voxels
$j\,\in\,\mathcal{N}(i)$ in the interrogation window.
This treatment would introduce a significant computational burden,
since at each location we would have to compute the gradient for all
$|\mathcal{N}(i)|$ voxels in the neighborhood.
A computationally more efficient procedure is to expand \eref{eq:ssd}
around the current flow estimate $\mathbf{v}_{0,j}$ for each voxel $j$.
After multiplying with $\lambda$ we arrive at our convexified data term:
\begin{equation}
\begin{aligned}
G_{\operatorname*{SSD},U_0,U_1}(i,&\mathbf{v}_i) = \lambda\!\!\sum_{j\in\mathcal{N}(i)} |U_0(j)-U_1(j+\mathbf{v}_{0,j})
- \\&
 (\mathbf{v}_i-\mathbf{v}_{0,j})\trans \nabla U_{1|(j+\mathbf{v}_{0,j})}|^2 \omega(j-i).
\end{aligned}
\end{equation}
With this formulation we only have to evaluate the gradients and volumes
once per voxel, namely at the current flow estimate.
The proximal map for the SSD at voxel $i$ again amounts to solving
a small quadratic problem to find the update for $\mathbf{v}_i$: 
\begin{equation}
\begin{aligned}
 \left((I +  \tau \partial G)^{-1}(\hat{\mathbf{v}})\right)_i =&
 \operatorname*{arg\,min}_{\mathbf{v}_i}
\frac{1}{2\tau} (\hat{\mathbf{v}}_i - \mathbf{v}_i)^2
+ \\&
G_{\operatorname*{SSD},U_0,U_1}(i,\mathbf{v}_i).
\end{aligned}
\end{equation}

As already mentioned, we embed the optimization of \eref{eq:energy}
into a coarse-to-fine scheme and adjust the data term by moving the
particles. Depending on the representation this step is often called
warping \cite{bro-04}.
After warping, the convex approximation of the data term is computed
and several iterations of \eref{eq:pd_iterates} are applied to
find an update of the flow estimate.
The coarse-to-fine scheme can easily be implemented by increasing the
radii in the descriptor.
To perform warping we need to rebuild the KD-tree for the first
time-step, moving the particles based on the current flow estimate.
Their motion $\mathbf{v}(p_i)$ is found by bilinear interpolation of
the flow at grid locations, which is in accordance with our
discretization of the operators $\nabla$ and $\div$.
Note that the chosen optimization scheme \cite{cha-11} is rather
generic and allows us to test different regularization and data terms.
Apart from allowing for acceleration, the smooth (and strongly convex)
soft-constraint regularizer \eref{eq:regularizers_QRDA} removes the
need for a second forward-backward step, and with that the need to
store any dual variables at all. E.g., one can employ the \emph{fast iterative shrinkage-thresholding algorithm}
(FISTA)~\cite{Beck2009} to perform the optimization.
We point out that the descriptor scheme implies an overall memory
footprint about an order of magnitude smaller than any algorithm,
which requires to explicitly store the dense voxel
volume, \eg, \cite{las-17}.

\section{Evaluation}\label{sec:evaluation}
\def\tablecaptionspace{-1em}

 \begin{figure*}[tb]
 \centering
 \includegraphics[width=0.28\textwidth]{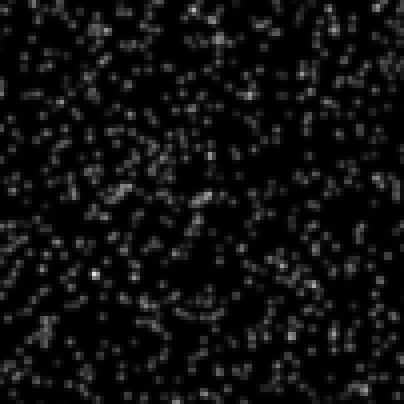}
 \includegraphics[width=0.28\textwidth]{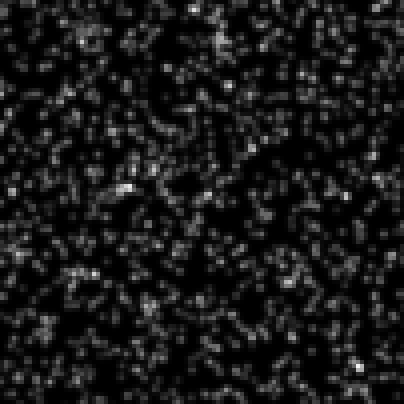}
 \includegraphics[width=0.28\textwidth]{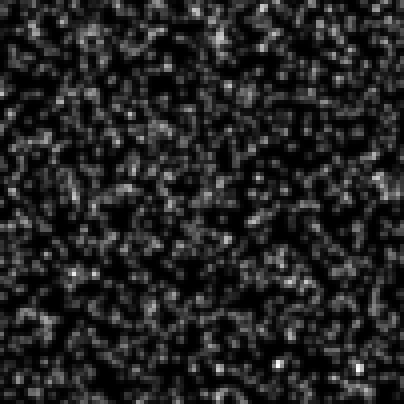}
 \caption{
Detail of rendered 2D particle images for particles densities of 0.075, 0.125
and 0.175 (from left to right).
 }
 \label{fig:ppp}
\end{figure*}

To quantitatively evaluate our approach and to compare different parameter
settings we have generated synthetic data using the Johns Hopkins turbulence
database (JHTDB)~\cite{li-08,per-07}. The database provides a direct numerical
simulation of isotropic turbulent flow in incompressible fluids.
We can sample particles at arbitrary positions in the volume and track them
over time. Additionally, we sample ground truth velocity flow vectors on a
regular voxel grid.

Our setup consists of four cameras and particles rendered into a volume of
$1024 \times 512 \times 352$ voxels. The cameras are arranged symmetrically
with viewing angles of $\pm 35^{\circ}$ w.r.t.~the $yz$-plane of the
volume and $\pm 18^{\circ}$ w.r.t.~the $xz$-plane.
Camera images have a resolution of $1500 \times 800$ pixels,
where a pixel is approximately the size of a voxel.
In this evaluation the maximum magnitude between two time steps is 4.8 voxels.

3D particles are rendered to the image planes as Gaussian blobs
with $\sigma=1$ and intensity varying between 0.3 and 1,
where the same intensity is used in all four cameras.
We test different particle densities from $0.075$ to $0.2$
particles per pixel (ppp).
A density of $0.1$ leads to a density of $\approx\frac{3}{10000}$ particles
per voxel.
Exemplary renderings for particle densities of $0.075$, $0.125$ and $0.175$
are shown in \fref{fig:ppp}.

\begin{figure*}[tb]
 \centering
 \includegraphics[width=0.325\textwidth]{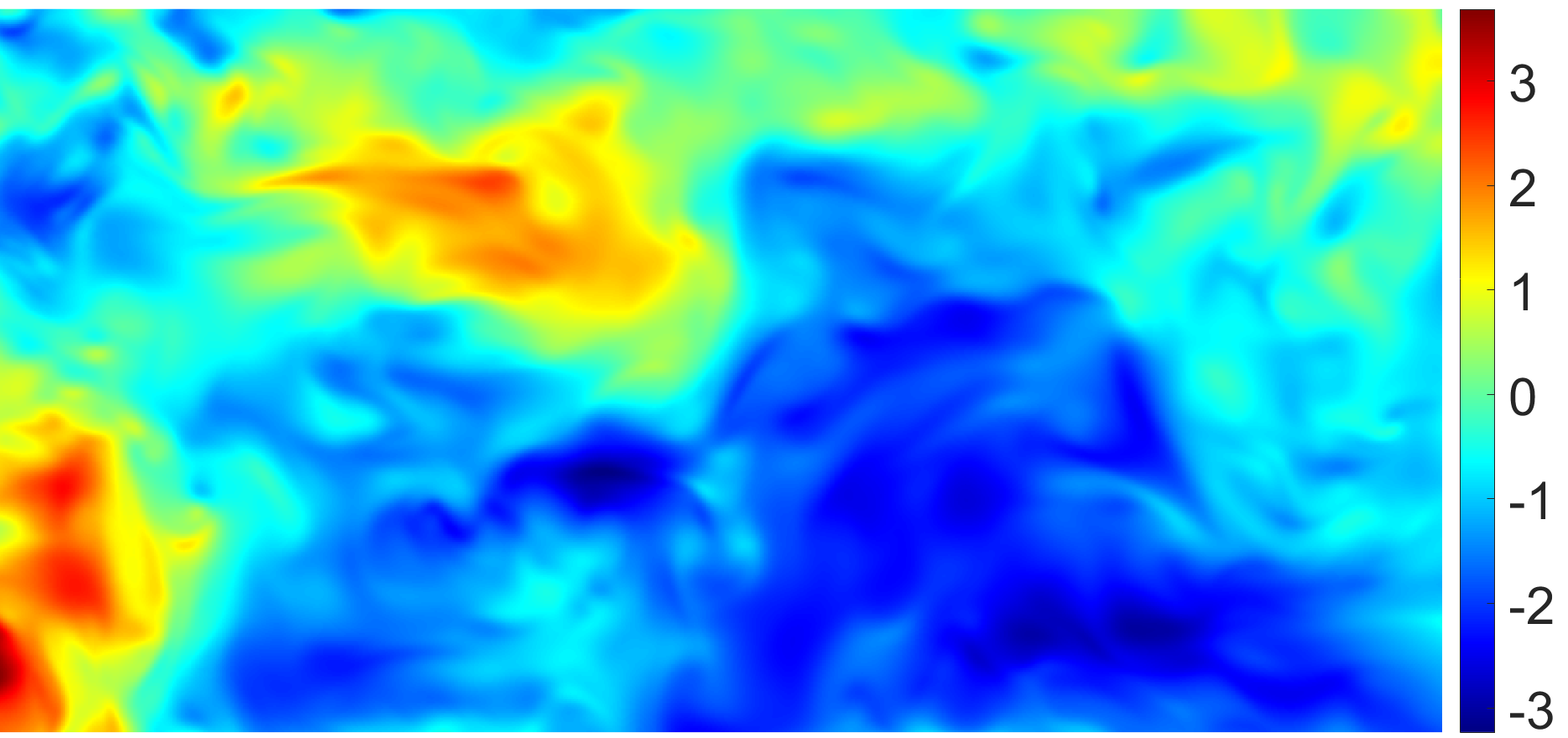}
 \includegraphics[width=0.325\textwidth]{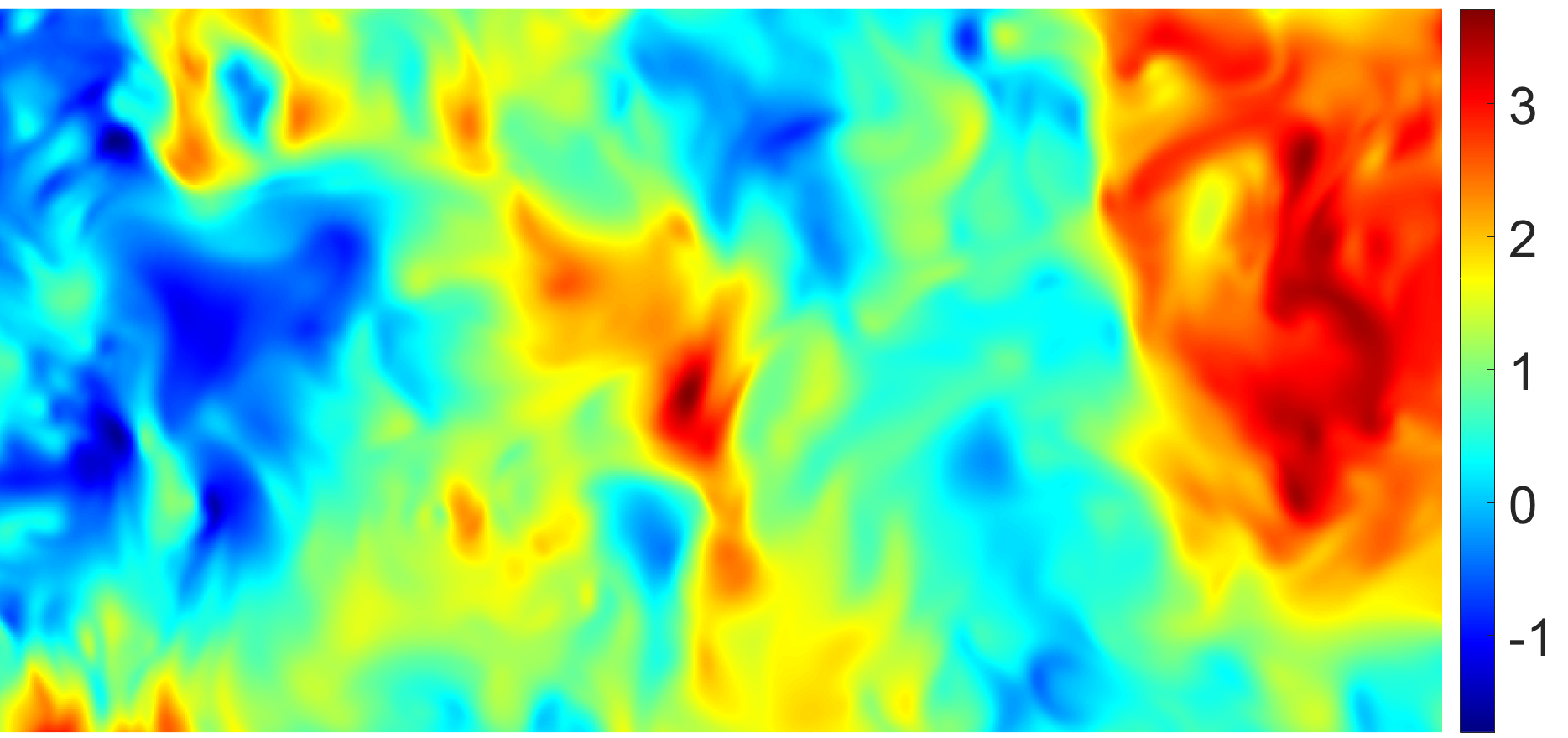}
 \includegraphics[width=0.325\textwidth]{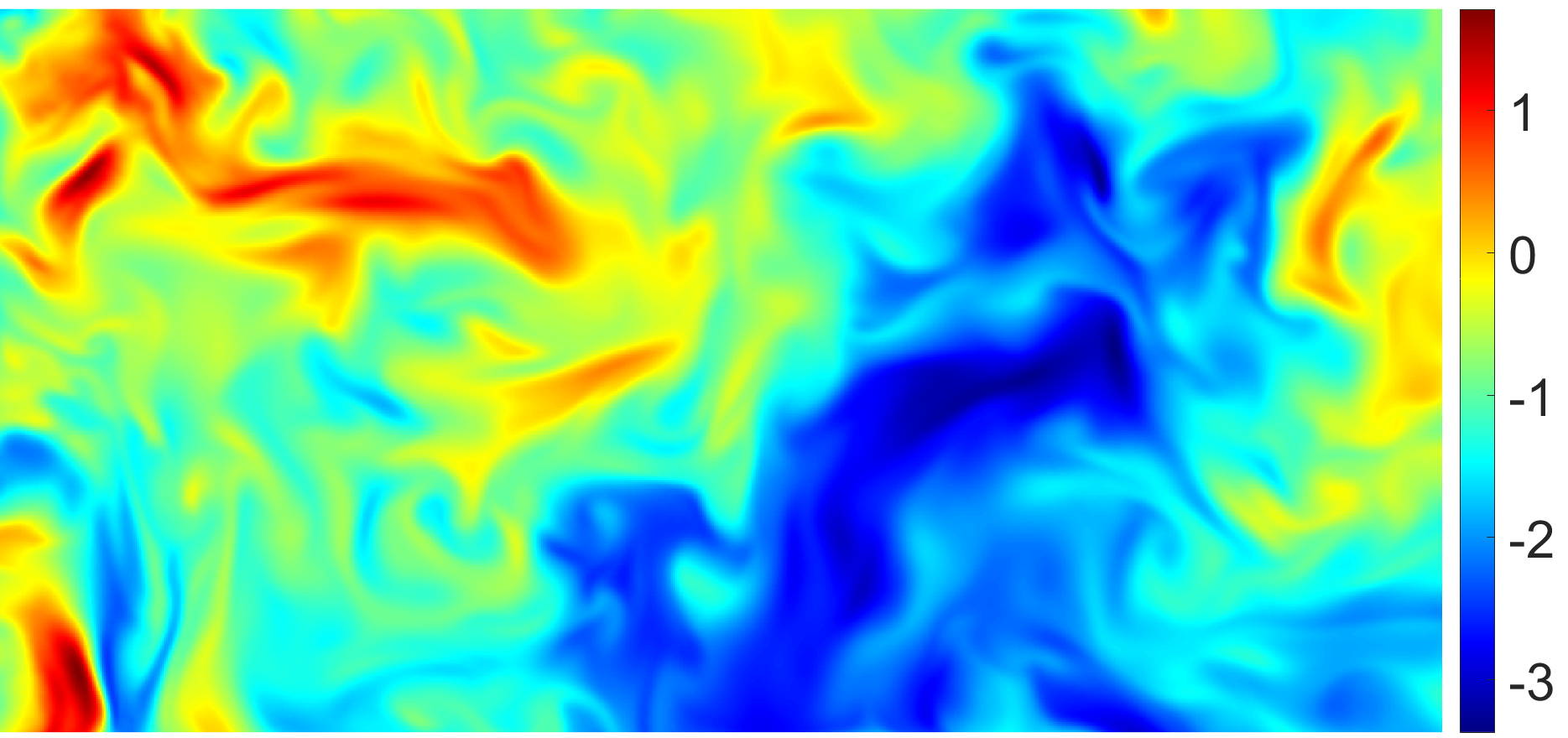} \\
 \includegraphics[width=0.325\textwidth]{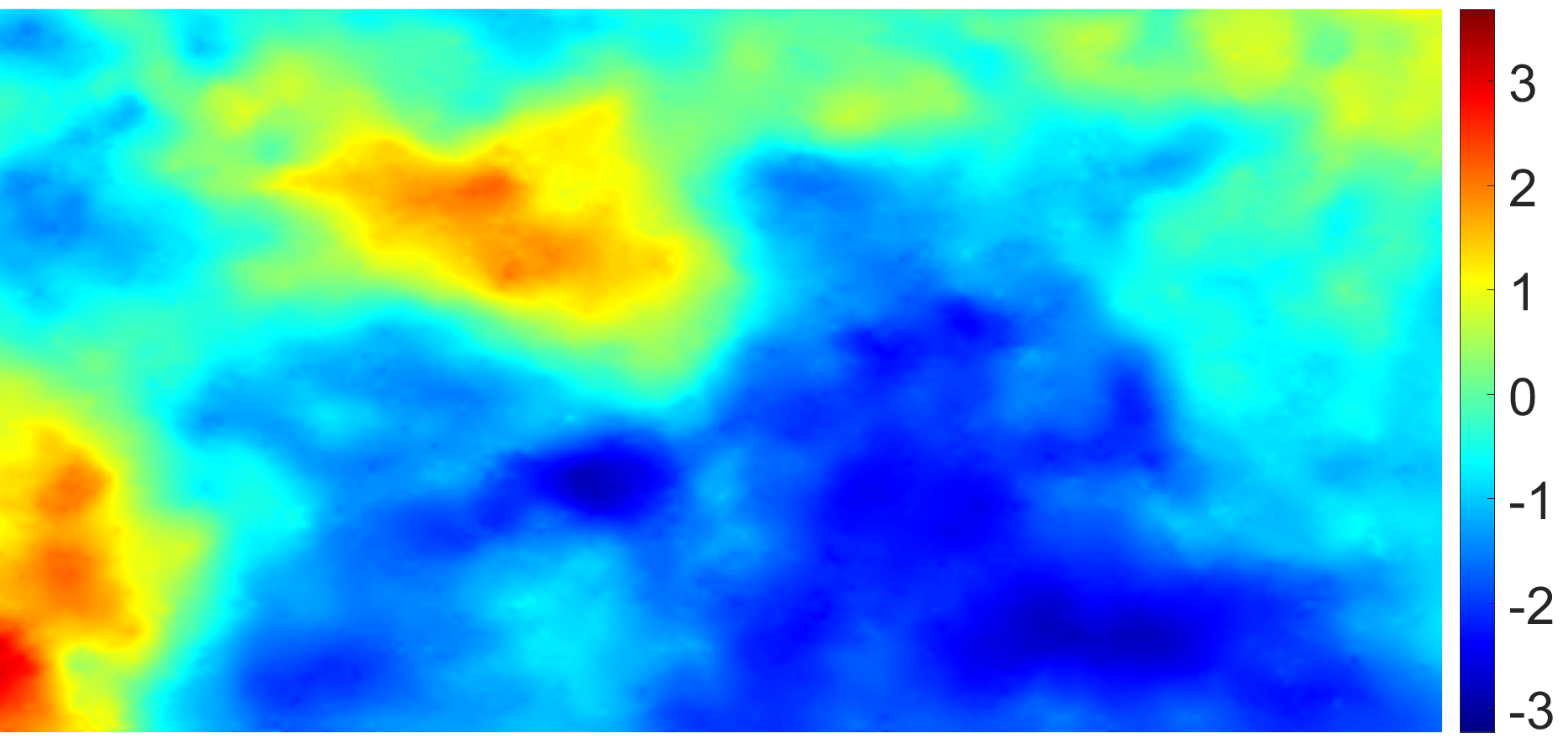}
 \includegraphics[width=0.325\textwidth]{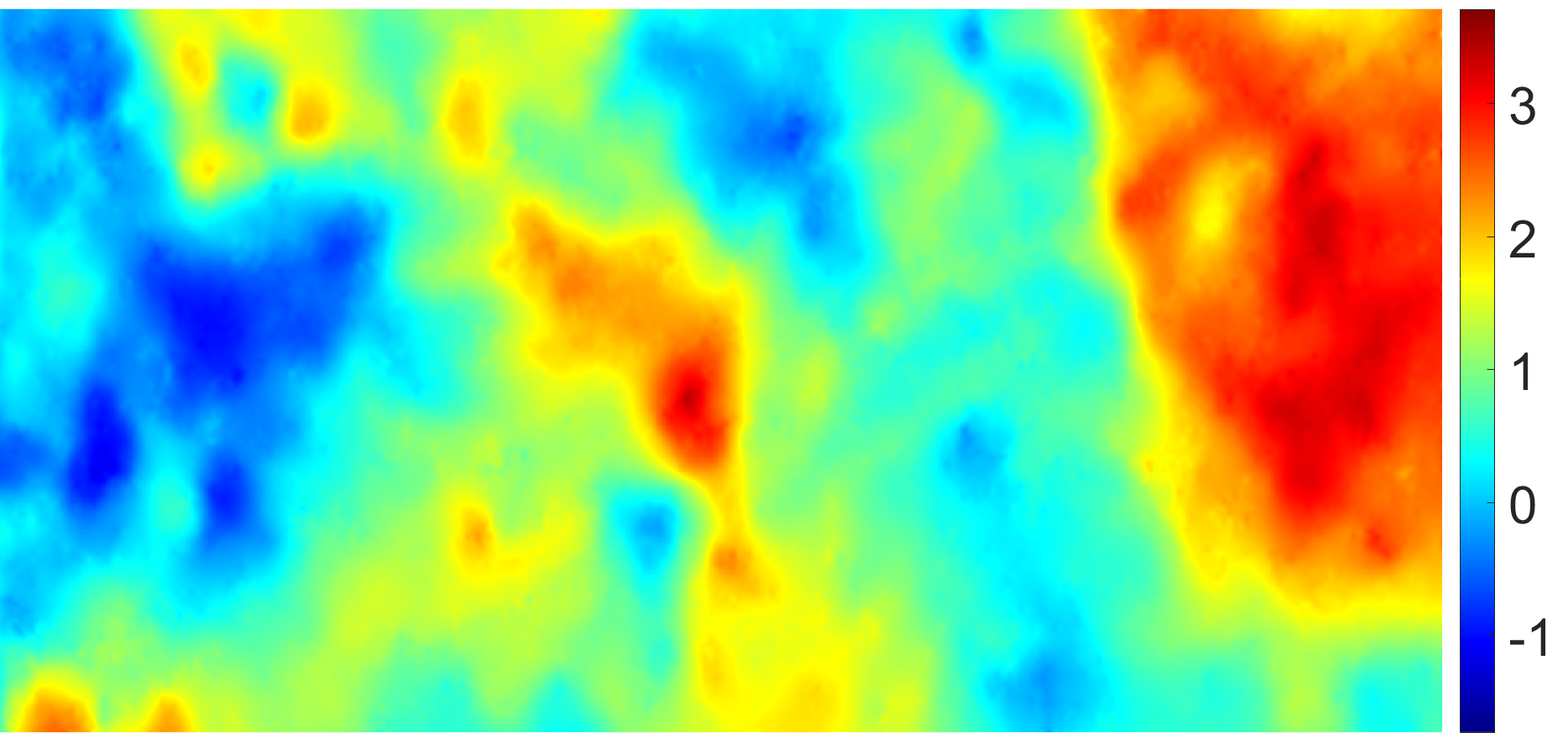}
 \includegraphics[width=0.325\textwidth]{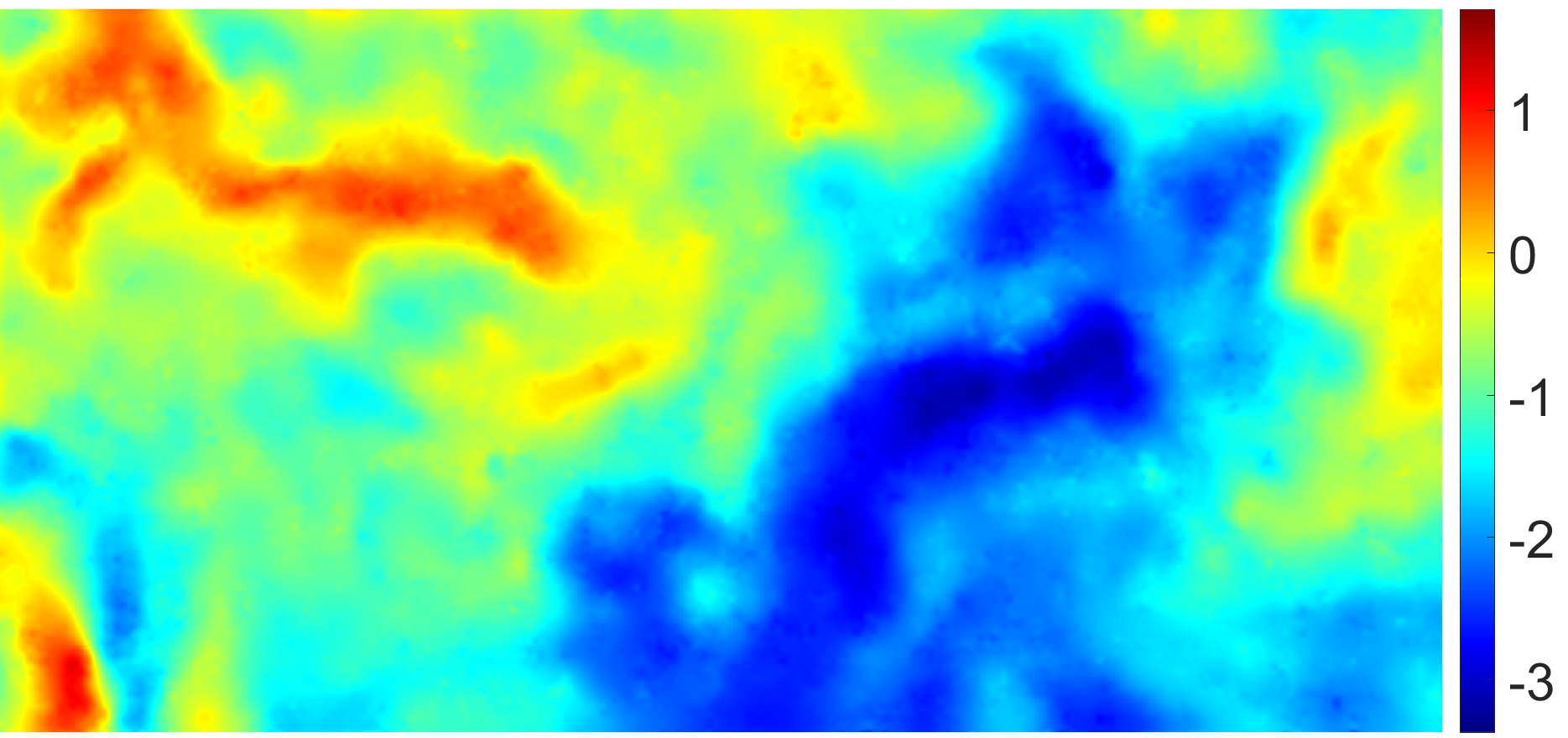}
 \caption{
\textit{xy}-slice of the flow in X-, Y- and Z-direction (left to right).
Top: Ground truth. Bottom: Estimated flow for $ppp=0.125$.
 }
 \label{fig:results}
\end{figure*}

We compare our method to~\cite{las-17} and test different particle densities
and parameter settings. Our main evaluation criterion is the
\textit{average endpoint error} (AEE) of our estimated flow vectors,
measured in voxels \wrt to the resolution of our test volume ($1024 \times 512 \times 352$).
Standard parameter settings for our coarse-to-fine scheme are 9 pyramid levels and a pyramid scale
factor of $0.95$ between adjacent pyramid levels.
Note that larger velocity magnitudes would require more pyramid levels.
We chose 10 warps and 20 inner iterations per pyramid level.
Unless otherwise specified we use SSD as data cost.
We also select our physically based regularizer
$\operatorname{QRD_\infty}$ for our quantitative evaluation.
According to \cite{las-17}, quadratic penalizing the gradient vectors of the flow,
while enforcing a divergence-free vector field, performed better than other
tested regularizers.
For IPR we chose an intensity threshold of $\theta=0.04$, inner iterations
$n_{steps}=40$ and 24 triangulation iterations, starting from a triangulation
error $\epsilon=0.8$ and increasing it up to $\epsilon=2.0$.
Note that for lower particle densities less iterations would already be sufficient.
However, most particles will be triangulated already in earlier iterations,
such that the additional runtime demand for more iterations becomes negligible.
Flow vectors are estimated on 
every fourth voxel location per dimension,
\ie~a $256 \times 128 \times 88$ voxel grid, and subsequently upscaled to
$1024 \times 512 \times 352$ in order to be compared with the ground truth
flow field.
As shown in~\cite{las-17} flow estimation on the down-sampled grid can be
performed with virtually no loss in accuracy, but with the benefit of faster
computation and lower memory demands.
In \fref{fig:results} we visualize a single \textit{xy}-slice (z=72) of the
flow field for both, ground truth and our estimated flow, with standard
settings at a particle density of $ppp=0.125$.

In addition to the quantitative evaluation we show qualitative results on test
case D of the $4^{th}$ International PIV Challenge~\cite{kah-16}.

\subsection{3D Reconstruction \& Particle Density}

We start by comparing different initialization methods for various particle
densities.
Additional to our implementations of MART and IPR we also provide results
for starting the flow reconstruction from the ground truth particle
distributions. Following the naming convention of~\cite{kah-16},
we refer to this version as HACKER.
In \tref{tab:initMethod} we compare the estimated flow \wrt the AEE and
furthermore show statistics on the particle reconstruction of time step 0,
for both reconstruction techniques, MART and IPR.
Reconstructed particles with no ground truth particle within a distance
of 2 voxels are considered as ghost particles.
As expected, assuming perfect particle reconstruction (HACKER) the flow
estimation is still improving with higher particle densities.
However, for increasing particle densities the reconstruction gets more
ambiguous, as can be seen from the increased number of ghost particles,
as well as missing particles for both MART and IPR.
The numbers suggest that the current bottleneck of our pipeline can be
found in the particle reconstruction.
Furthermore, IPR consistently outperforms MART at all particle densities.
Here, MART already starts to fail at lower particle densities while our IPR
implementation yields results similar to HACKER up to a particle density
of $0.15$.

\begin{table*}
\caption{\label{tab:initMethod} Comparison of different initialization
methods (MART vs. IPR vs. HACKER).}
\vspace{\tablecaptionspace}
\setlength\tabcolsep{0.3cm}
\begin{center}
\begin{tabular}{@{}llllllll}
\hline
\rowcolor{gray!10}
\multicolumn{2}{l}{Particle Density (ppp)} & 0.075 & 0.1 & 0.125 & 0.15 & 0.175 & 0.2 \\
\hline
MART & Avg.~endpoint error & 0.2768 & 0.2817 & 0.2927 & 0.3119 & 0.3363 & 0.3500 \\
& Undetected particles & 1412 & 6283 & 13918 & 24141 & 37215 &  51190 \\
& & 3.21\% & 10.7\% & 18.96\% & 27.41\% & 36.22\% & 43.59\% \\
& Ghost particles & 22852 & 44752 & 79356 & 107276 & 118476 & 129791 \\
& & 51.89\% & 76.21\% & 108.11\% & 121.79\% & 115.29\% & 110.52\% \\
& Avg.~position error & 1.681 & 1.672 & 1.663 & 1.654 & 1.647 & 1.638 \\
\hline
IPR & Avg.~endpoint error & 0.2505 & 0.2388 & 0.2250 & 0.2260 & 0.2522 & 0.2950 \\
& Undetected particles & 18 & 24 & 35 & 582 & 14187 & 33936 \\
& & 0.04\% & 0.04\% & 0.05\% & 0.66\% & 13.81\% & 28.90\% \\
& Ghost particles & 4 & 3 & 14 & 4344 & 107286 & 176176 \\
& & 0.01\% & 0.01\% & 0.02\% & 4.93\% & 104.40\% & 150.01\% \\
& Avg.~position error & 0.001 & 0.001 & 0.002 & 0.029 & 0.188 & 0.301 \\
\hline
HACKER & Avg.~endpoint error & 0.2503 & 0.2382 & 0.2244 & 0.2229 & 0.2140 & 0.2068 \\
\hline
\end{tabular}
\end{center}
\end{table*}

\subsection{Descriptor \& Data Cost}

We evaluate two different distance metrics for our sparse descriptor:
\textit{sum of squared distances} (SSD) and
(negative) \textit{normalized cross-correlation} (NCC).
In addition to our spherical-shaped descriptor introduced in
\sref{sec:sparseDescriptor}, we evaluate our approach with a descriptor based
on a regular grid structure.
A grid point is set at every second voxel position per dimension (\ie~-2,0,2)
within a radius of 8, resulting in 257 grid points.
Particles within a radius of $d=10.5$ are considered for the descriptor estimation.
Here, we splat particles to $k=8$ regular grid points, and utilize $k=5$ for our
standard layer based descriptor
(with denser sampling closer to the descriptor center).
The impact of these choices on the estimated flow vectors
are summarized in \tref{tab:descriptor} for various particle densities.
Here both distance metrics, SSD and NCC, perform similarly well.
Our spherical, layer based descriptor appears to work slightly,
but also consistently, better than the descriptor based on the regular grid.

\begin{table*}
\caption{\label{tab:descriptor} Comparison of average endpoint error for
different descriptors with IPR initialization.}
\vspace{\tablecaptionspace}
\setlength\tabcolsep{0.5cm}
\begin{center}
\begin{tabular}{@{}llllll}
\hline
\rowcolor{gray!10}
\multicolumn{2}{l}{Particle Density (ppp)} & 0.075 & 0.1 & 0.125 & 0.15 \\
\hline
NCC & Layered spherical descriptor (331, d=21) & 0.2492 & 0.2359 & 0.2270 & 0.2286 \\\ 
 & Regular spherical grid (257, d=21) & 0.2630 & 0.2498 & 0.2405 & 0.2431 \\

\hline
SSD &  Layered spherical descriptor (331, d=21) & 0.2505 & 0.2388 & 0.2250 & 0.2260 \\ 
 & Regular spherical grid (257, d=21) & 0.2643  & 0.2436 & 0.2362  & 0.2374    \\

\hline
\end{tabular}
\end{center}
\end{table*}

\subsection{Comparison to Dense VolumeFlow}

We compare our method to~\cite{las-17} where the data term is based on a
high-resolution voxel grid, as obtained from MART.
Additionally to MART reconstruction we also show results for IPR and HACKER,
where particles were rendered into the volume by tri-linear interpolation to
their nearest $2 \times 2 \times 2$ voxels.
For the flow estimation we chose the recommended parameter set
of~\cite{las-17}: SSD with IV size $13^3$, $\operatorname{QRD_\infty}$ and stepsize $h=4$.
Average endpoint errors are shown in ~\tref{tab:dense} for different
particle densities.
Performance is slightly better than for our sparse descriptor variant,
with the drawback that a memory demanding high resolution intensity volume is
needed for the inherently sparse data.
By adding more grid points to our descriptor
(i.e.~a grid point at every voxel location) we could effectively rebuild a
dense matching approach with our sparse data.
However, we opt for a more compact representation that allows faster
computation and less memory requirements.
Furthermore, our sparse descriptor can be adapted to any kind of
similarity measure, where the use of numerical differentiation allows for a
simple and quick evaluation.
Here, it would be interesting to test also learned similarity functions,
which could benefit from our compact representation~\cite{Yi16,zbo-16,kho-17}.

\begin{table}
\caption{\label{tab:dense} Average endpoint error of dense method for different initialization methods (MART vs. IPR vs. HACKER).}
\vspace{\tablecaptionspace}
\setlength\tabcolsep{0.16cm}
\begin{center}
\begin{tabular}{@{}lllll}
\hline
\rowcolor{gray!10}
Particle Density (ppp) & 0.075 & 0.1 & 0.125 & 0.15 \\ 
\hline
MART  & 0.2470 & 0.2717 & 0.3144 & 0.3735 \\ 
IPR  & 0.2266 & 0.2107  & 0.2001 & 0.2057 \\ 
HACKER & 0.2270 & 0.2112 & 0.2000 & 0.1923 \\ 

\hline
\end{tabular}
\end{center}
\end{table}

\subsection{PIV Challenge Data}

\begin{figure}[tb]
 \centering
  \includegraphics[width=\columnwidth]{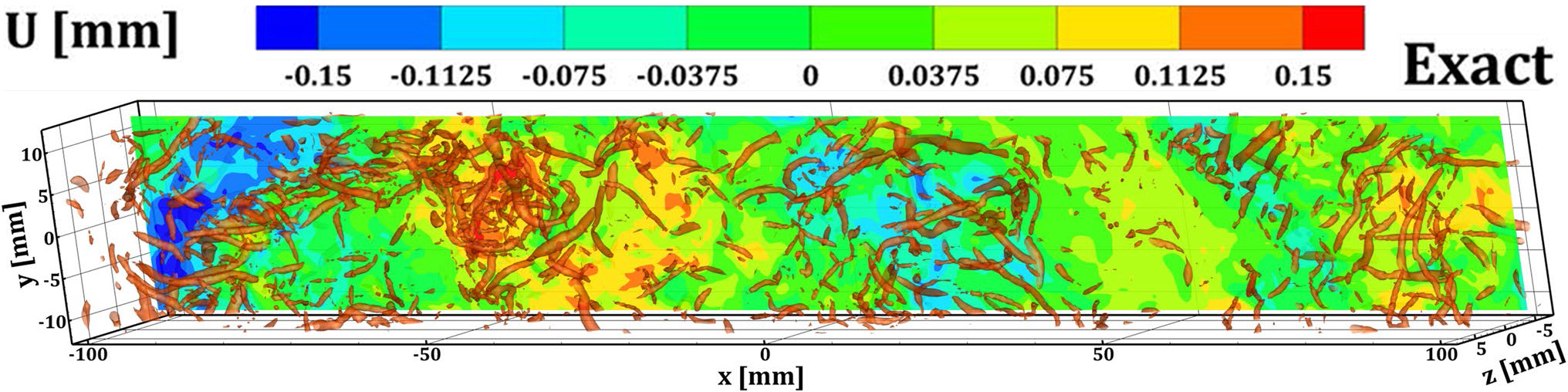} \\
  \includegraphics[width=\columnwidth]{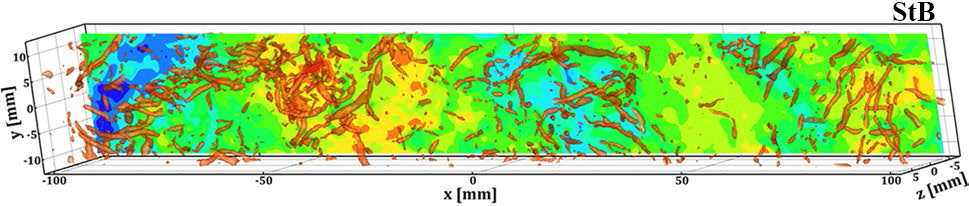} \\
 \includegraphics[width=\columnwidth]{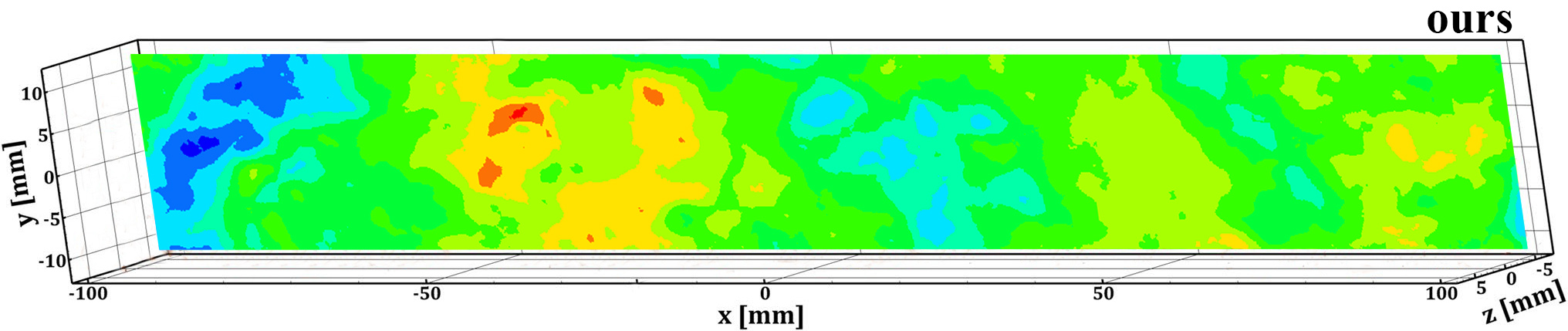} 
 \caption{
Visualization of xy-slice (z=-5.2mm) of snapshot 23 of the 4th International
PIV Challenge~\cite{kah-16} with discrete color coding to visually compare to
results of participators of the challenge. \emph{Top:} Ground truth (with additional visualization of vortices). \emph{Middle:} StB~\cite{sch-16}, the best performing method of the challenge. \emph{Bottom:} Our result. Images of ground truth and StB reproduced from~\cite{kah-16}.
 }
 \label{fig:pivchall}
\end{figure}

Test case D of the $4^{th}$ International PIV Challenge~\cite{kah-16} consists
of 50 time steps of a volume of size $4096 \times 512 \times 352$ and a
particle density of $0.1$.
Unfortunately no ground truth is provided, however, we can visually compare our
results to results stated in the challenge.
In \fref{fig:pivchall} we show result of our method for snapshot 23 of the challenge, as well as the ground truth and the result of the top-performing method \emph{Shake-the-box} (StB)~\cite{sch-16}. The motion fields appear comparable, flow details are recovered.
Note that our method uses only two frames to estimate the motion field, while StB relies on multiple time steps.

\section{Conclusion}
Our method is based on and extends work of a conference publication
\cite{las-17}.
In particular, we adopt their physically motivated regularization scheme
and follow the same optimization methodology.
We extend \cite{las-17} with a novel, energy based IPR method that allows for
more accurate and robust particle reconstructions. 
Relying on the quality of the 3D particle reconstruction,
these advantages also affect the accuracy of our motion estimation technique,
which before was based on the tomographic reconstruction model of MART.
To further exploit sparsity of the directly reconstructed particles in 
the volume, we propose a novel sparse descriptor that can be constructed 
on-the-fly and possess a significantly lower memory footprint 
than a dense intensity volume, which is required in high resolution 
to deliver motion reconstructions of similar quality. 

A current draw-back of our model is that the initially reconstructed 
particles are kept fixed and the mutual dependency of fluid motion and
particle reconstruction is not considered. 
Hence, an interesting path to investigate in the future would be to combine 
particle reconstruction and flow estimation in a joint energy model. 

\section*{Acknowledgments}
This work was supported by ETH grant 29~14-1.  Christoph Vogel and Thomas Pock
acknowledge support from the ERC starting grant 640156, ’HOMOVIS’.
We thank Markus Holzner for discussions and help with the
evaluation.

{\small
\bibliographystyle{ieeetr}
\bibliography{references}
}

\end{document}